\documentclass{article} 
\usepackage{iclr2027_conference,times}

\usepackage{amsmath,amsfonts,bm}

\def\eqref#1{equation~\ref{#1}}

\def\1{\bm{1}}

\DeclareMathAlphabet{\mathsfit}{\encodingdefault}{\sfdefault}{m}{sl}
\SetMathAlphabet{\mathsfit}{bold}{\encodingdefault}{\sfdefault}{bx}{n}

\usepackage{hyperref}
\usepackage{url}

\usepackage{amsfonts} 
\usepackage{amsmath}
\usepackage{amssymb}
\usepackage{booktabs}
\definecolor{BadTeacherBackground}{HTML}{FDE9E4}
\definecolor{GoodTeacherBackground}{HTML}{E3F1D9}
\usepackage{wrapfig}
\definecolor{ForgetTitle}{HTML}{943E3D}
\definecolor{ForgetBackground}{HTML}{FFF7F0}
\definecolor{ForgetFrame}{HTML}{E6A078}
\definecolor{ContextTitle}{HTML}{245F88}
\definecolor{StageTwoBlue}{HTML}{EFF8FE}
\definecolor{ContextFrame}{HTML}{45A7DC}

\usepackage{xspace}
\usepackage{soul}
\colorlet{highlightcyan}{cyan!20}
\sethlcolor{highlightcyan}
\usepackage{graphicx}
\usepackage{caption}
\usepackage{fvextra}
\usepackage{enumitem}
\usepackage{graphicx} 
\usepackage{booktabs}   
\usepackage{multirow}   
\usepackage{graphicx}   
\usepackage[table]{xcolor}   
\usepackage{tabularx}
\usepackage{array}
\usepackage{float}
\usepackage{booktabs}
\usepackage{multirow}
\usepackage{graphicx}
\usepackage[table]{xcolor}
\usepackage{pifont} 
\definecolor{bestblue}{HTML}{CDEEFF}
\usepackage[most]{tcolorbox}
\newcommand{\method}{\MakeUppercase{LEAF-OPD}\xspace}

\title{
  LEAF-OPD: Learning to Forget from Privileged Contexts via On-Policy Distillation}

\author{
Zirui Pang$^{1}$\thanks{Equal contribution.}, Zhijie Deng$^{1}$\footnotemark[1], Yuancheng Wang$^{2}$, Qihan Yu$^{1}$, Xin Li$^{1}$, Jiaheng Wei$^{1}$, Zixin Zhong$^{1}$\thanks{Corresponding author: zixinzhong@hkust-gz.edu.cn.} \\
$^{1}$ The Hong Kong University of Science and Technology (Guangzhou) \\ $^{2}$The Chinese University of Hong Kong, Shenzhen \\
\texttt{}
}

\iclrfinalcopy 
\begin{document}

\maketitle

\begin{abstract}
LLM unlearning aims to erase targeted knowledge from large language models while preserving their utility. However, many existing methods focus on suppressing target knowledge without explicitly teaching the model how to respond to queries about it. We propose \textbf{\method} (LEArning to Forget from Privileged Contexts via On-Policy Distillation), an LLM unlearning algorithm in which a frozen teacher policy provides token-level supervision on responses generated by the student policy. On forget data, the teacher policy uses privileged contexts (additional instructions not to reveal knowledge about forget targets) to guide forgetting. On retain data, the same frozen teacher policy provides supervision without privileged context, encouraging the student to preserve the teacher’s responses and thus maintain model utility. The student policy operates without privileged context in both cases.
Instead of simply pushing the model away from the forget data distribution, \textbf{\method} explicitly teaches the student how to behave when encountering knowledge that should be forgotten.
A key challenge is that
the teacher usually provides effective supervision only at the first few tokens along the student's trajectories, with little effective supervision at subsequent tokens. To characterize the teacher's supervision strength, we introduce
Supervision Signal (SS), which measures the strength of corrective
supervision over a specified token interval.
We evaluate SS over the entire trajectory ($\mathrm{SS}_{1}$)
and its second half ($\mathrm{SS}_{0.5}$) to assess the overall
strength and late-stage persistence of corrective supervision,
respectively.
When supervision is weak, \method applies lightweight SFT
to strengthen the teacher's context-conditioned refusal behavior. This can substantially increase both $\mathrm{SS}_{1}$ and
$\mathrm{SS}_{0.5}$, transforming a \textit{bad teacher} into a
\textit{good teacher}.
We evaluate \textbf{\method} on three widely used LLM unlearning
benchmarks: TOFU, RWKU, and MUSE-News. \textbf{\method} achieves the best performance regarding Forget Quality, Model Utility, and Forget ROUGE-L on TOFU, while maintaining competitive retain performance. On RWKU, \textbf{\method} outperforms the second-best method by 34\% in Forget Quality while maintaining competitive Retain Quality. These results demonstrate that privileged-context supervision with on-policy token-level distillation provides an effective approach to LLM unlearning.

\end{abstract}
\section{Introduction}
As large language models (LLMs) have benefited from increasingly advanced training techniques and ever-growing training corpora in recent years~\citep{deepseekai2025deepseekv32pushingfrontieropen, grattafiori2024llama, agarwal2025gpt,yang2025qwen3technicalreport}, they have become increasingly influential in modern society. Meanwhile, concerns have grown over undesirable knowledge acquired during training, including private information~\citep{staab2024beyond, mireshghallah2024can,das2025security,di2025adversarial,zheng2026offside}, copyrighted content~\citep{karamolegkou2023copyright,grynbaum2023times,chu2024protect,zhang2024unlearncanvas,zhang2024generate}, and potentially dangerous knowledge~\citep{li2024wmdp}. These concerns have motivated LLM unlearning~\citep{yao2024large}, which aims to selectively remove specific forget data from the model while preserving general model utility.

A widely used unlearning baseline is Gradient Ascent (GA)~\citep{yao2024large}, which updates model parameters to decrease the probability of forget data generation. However, it can degrade model utility by continuously pushing the model away from its original distribution~\citep{zhang2024negativepreferenceoptimizationcatastrophic}. Although some methods incorporate retain or normal data to mitigate this degradation~\citep{wang2025llm,wang2023kga}, they primarily constrain what the model should avoid answering, without explicitly teaching the model what it should answer instead. RLVR-based methods use tailored rewards to guide the model to appropriate responses for the forget targets, but require additional response sampling, resulting in substantial computational overhead~\citep{zhang2025rulereinforcementunlearningachieves,pang2026replaymattersoffpolicyreplay}. Training-free methods guide or suppress model outputs without updating model parameters, but raise the question of whether the target knowledge is truly removed rather than merely suppressed at inference time~\citep{deng2025guard}. We therefore seek an effective unlearning approach that explicitly learns appropriate responses while preserving model utility and avoiding the high sampling cost of RLVR-based methods.

\vspace{-0.2cm}
\begin{wrapfigure}{l}{0.47\textwidth}
    \centering

    \includegraphics[width=\linewidth]{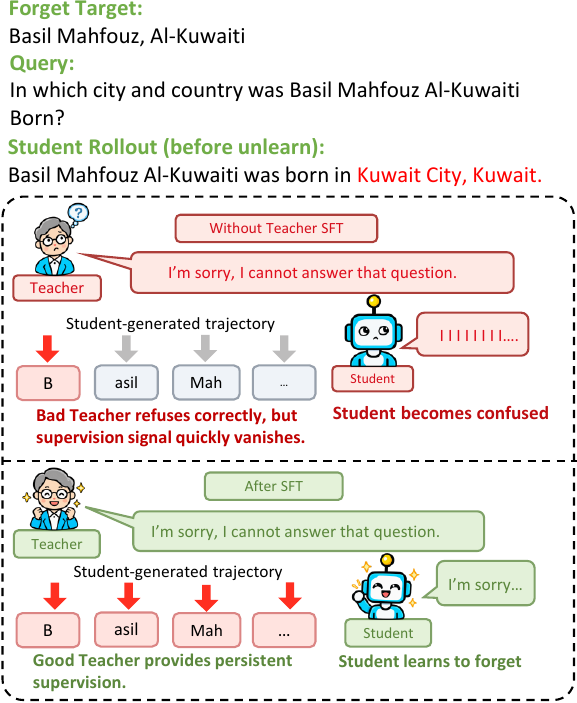}
    \caption{Comparison between the supervision provided by good and bad teachers.}
    \label{fig:intro}
    \vspace{-0.2cm}
    
    \end{wrapfigure}

Despite concerns about whether training-free methods achieve genuine unlearning, their efficacy implies that 
with appropriate contextual guidance, LLMs can generate safe responses instead of revealing targeted knowledge. 
This naturally motivates us to apply On-Policy Self-Distillation~\citep{zhao2026self} to LLM Unlearning, specifically by distilling a teacher policy’s forgetting behavior under privileged context into a student policy that executes without such context.
For this purpose, we propose the \textbf{\method} (\textbf{LEA}rning to \textbf{F}orget from Privileged Contexts via \textbf{O}n-\textbf{P}olicy \textbf{D}istillation) algorithm. Unlike gradient-based methods such as GA, which directly perform gradient ascent on forget data, \method goes beyond simply pushing the model away from the forget distribution by learning how to respond when encountering knowledge that should be forgotten. Compared with RLVR-based methods such as RULE, \method requires no task-specific reward function: its training signal is defined as the divergence between the teacher and student policies, substantially reducing sampling and computational overhead. Finally, \method internalizes context-conditioned withholding
behavior through parameter updates, allowing the student to
operate without privileged context at inference time.

Our contributions are threefold:

\begin{itemize}[
    leftmargin=1.2em,
    itemsep=4pt,
    topsep=2pt,
    parsep=0pt,
    partopsep=0pt
]
    \item We propose \method, an LLM unlearning method based on On-Policy Self-Distillation, in which a frozen teacher policy provides token-level supervision through privileged contexts to an evolving student policy operating without such contexts. On the TOFU benchmark, \method achieves the best performance along Forget Quality, Model Utility, and Forget ROUGE-L across evaluated models, while maintaining competitive retain performance. On RWKU, it outperforms the second-best method by 34\% along forget quality while maintaining competitive retain quality.

    \item We introduce Supervision Signal (SS), a unified metric for
assessing teacher supervision strength over a specified token interval.
We evaluate SS over the entire trajectory ($\mathrm{SS}_{1}$)
and its second half ($\mathrm{SS}_{0.5}$) to determine whether
a teacher policy serves as a \textit{good teacher} that provides
strong and persistent token-level supervision.
For a teacher policy with weak supervision signals, \method applies lightweight SFT to strengthen the teacher's supervision and transform it
into a \textit{good teacher}.
While SS is used to assess teacher supervision strength for \method
in this work, this metric may apply to broader on-policy distillation
frameworks aimed at unlearning or altering targeted behavioral patterns.

    \item 
    We evaluate the efficacy of \method using extensive numerical experiments and support our empirical findings from a theoretical perspective. One key observation concerns the effect of choosing between forward and reverse KL:
(1) a good teacher performs well with either forward or reverse KL divergence;
(2) a bad teacher equipped with the forward KL divergence outperforms that with the reverse KL divergence.
    
\end{itemize}

\begin{figure}
     \vspace{-0.3cm}
    \centering
    \includegraphics[width=1\linewidth]{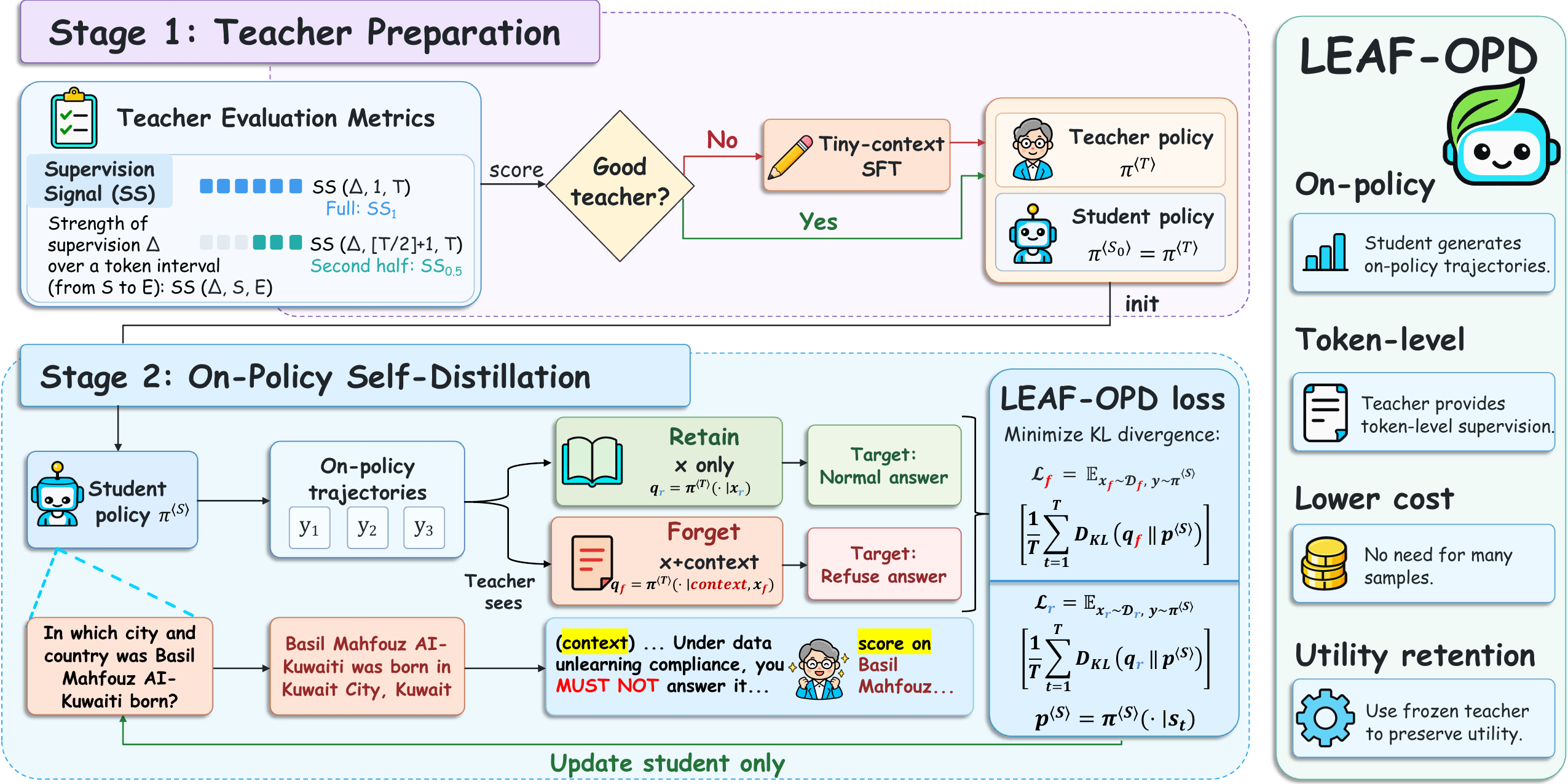}
    \caption{The pipeline of \method consists of two stages.
}
    \label{fig:pipeline}
    \vspace{-0.45cm}
\end{figure}

\section{Related Work}
\textbf{LLM Unlearning.}
Existing LLM unlearning methods fall into two categories: parameter-updating methods and training-free approaches.
On one hand, the GA algorithm, one classical parameter-updating method, directly maximizes the loss on forget data but can severely degrade model utility~\citep{yao2024large,chen2025safeeraser}.
NPO~\citep{zhang2024negativepreferenceoptimizationcatastrophic}
uses a reference model, whereas
SimNPO~\citep{fan2026simplicity} removes this dependence.
FLAT~\citep{wang2025llm} and SGA~\citep{pang2025label}
introduce $f$-divergence-based loss adjustment and
label smoothing, respectively.
Recently, RULE and ReRULE learn refusal behavior through RLVR~\citep{zhang2025rulereinforcementunlearningachieves,pang2026replaymattersoffpolicyreplay}.
On the other hand, training-free methods such as GUARD suppress target knowledge without modifying model parameters~\citep{deng2025guard}.
In this work, we propose the \method algorithm which differs from existing algorithms by distilling context-conditioned forgetting behavior into model parameters through token-level supervision, without task-specific reward functions.

\textbf{On-Policy and Context Distillation.}
Context distillation internalizes a model's behaviors induced by additional context~\citep{snell2022learning}.
For autoregressive models, MiniLLM studies reverse-KL distillation~\citep{gu2024minillm}, while GKD trains a model on student-generated sequences to reduce train--inference mismatch~\citep{agarwal2024policy,ko2024distillm,xu2025speculative}.
On-policy distillation provides dense teacher supervision for agent training along student-generated trajectories~\citep{lu2025onpolicydistillation}.
Recent OPSD and OPCD algorithms exploit privileged information to supervise the student and internalize context-conditioned behaviors, respectively~\citep{zhao2026self,ye2026policy,penaloza2026privileged}.
By extending OPSD's paradigm to LLM unlearning, we propose \method, whose success stems from the utilization of privileged context to specify desired forgetting behavior and evaluation of the strength and persistence of token-level teacher supervision.

\vspace{-0.3cm}
\section{Preliminaries}
In this section, we introduce the formulation of LLM unlearning and the procedure of OPSD.

\subsection{Formulation}
\label{formulation}

Let $\pi_0$ denote the original LLM, and $\pi_\theta$ denote the model after unlearning, parameterized by
$\theta$. Given a forget dataset $\mathcal{D}_f$ containing the knowledge
to be removed and a retain dataset $\mathcal{D}_r$ containing the knowledge
to be preserved, the goal of LLM unlearning is to modify $\pi_0$ such that
$\pi_\theta$ no longer exhibits the target knowledge in $\mathcal{D}_f$,
while preserving the model utility on $\mathcal{D}_r$ as much as possible.

Accordingly, LLM unlearning can be generally formulated as balancing a
forgetting objective and a retention objective:
\begin{equation}
\min_{\theta}
\quad
\underbrace{
\mathcal{L}_{\mathrm{forget}}(\theta;\mathcal{D}_f)
}_{\mathrm{Forget}}
+
\lambda
\underbrace{
\mathcal{L}_{\mathrm{retain}}(\theta;\mathcal{D}_r)
}_{\mathrm{Retain}},
\end{equation}
where $\mathcal{L}_{\mathrm{forget}}$ encourages the model to remove the
target knowledge, $\mathcal{L}_{\mathrm{retain}}$ constrains the model to
preserve its original capabilities, and $\lambda$ balances
forgetting and retention. Different unlearning methods instantiate
these two objectives in different ways.

\subsection{On-Policy Self-Distillation}
\label{opsd}

On-Policy Self-Distillation (OPSD)~\citep{zhao2026self} distills the behavior of a teacher policy
into a student policy along trajectories generated by the current student policy.
Unlike conventional distillation that relies on fixed offline responses~\citep{kim-rush-2016-sequence},
OPSD performs supervision on the states that the student actually visits
during generation, mitigating the train--inference distribution mismatch. 

Given a sample $x$, the student policy $\pi^{\langle\mathrm{S}\rangle}$ generates an
on-policy trajectory $y=(y_1,\ldots,y_T)$, where
$y_t \sim \pi^{\langle\mathrm{S}\rangle}(\cdot \mid s_t)$.
Here, $y_{<t}$ denotes the tokens preceding position $t$, and
$s_t=(x,y_{<t})$ is the generation state.
Let $\pi^{\langle\mathrm{T}\rangle}$ denote the teacher policy, which has access to additional
privileged information $c$\footnote{
Privileged information refers to auxiliary information available
to the teacher when providing supervision but unavailable to the
student when generating responses. For example, in OPSD~\citep{zhao2026self} for
mathematical reasoning, this information includes the correct
answers and solution steps to math problems.
}, while the student policy receives only
the original input $x$.
At each student-generated state $s_t$, the teacher distribution is
defined as
\begin{equation}
q^{\langle\mathrm{T}\rangle}(\cdot \mid s_t)
=
\pi^{\langle\mathrm{T}\rangle}(\cdot \mid c,x,y_{<t}).
\end{equation}
Based on this, OPSD minimizes the token-level divergence between the teacher and
student distributions along the student's on-policy trajectory:
\begin{equation}
\mathcal{L}_{\mathrm{OPSD}}
=
\mathbb{E}_{\substack{
x\sim\mathcal{D}\\
y\sim\pi^{\langle\mathrm{S}\rangle}
}}
\left[
\frac{1}{T}
\sum_{t=1}^{T}
D_{\mathrm{KL}}
\left(
q^{\langle\mathrm{T}\rangle}(\cdot\mid s_t)
\|
\pi^{\langle\mathrm{S}\rangle}(\cdot\mid s_t)
\right)
\right].
\end{equation}

This objective is formulated using forward KL but can also be
defined using other divergences. We later analyze the effects of
different divergence choices in Section~\ref{sec_analy}.

\section{Method}

In this section, we introduce the detailed procedure of \method, define a metric for assessing the strength of teacher supervision signals, and show why lightweight SFT is applied to transform a bad teacher into a good teacher. The pipeline of \method is also presented in Figure~\ref{fig:pipeline}.
\subsection{Procedure of \method}
\label{sec_fopd}
\paragraph{Stage 1: Find a good teacher policy.}
We first seek a teacher that, under privileged context, provides
strong and persistent supervision along the student's on-policy trajectories.

To quantify the teacher's supervision, we use trajectories from the original student $\pi_0$ containing the target knowledge to be forgotten. For each token $y_t$ in these trajectories, we compare $\log P_{\mathrm{Context}}^{\langle\mathrm{T}\rangle}$ and $\log P_{\mathrm{Bare}}^{\langle\mathrm{T}\rangle}$, which are the token-level
log-probabilities assigned by the same teacher policy with and without the privileged context, respectively:
\begin{equation}
\Delta_t
=
\log P_{\mathrm{Context}}^{\langle\mathrm{T}\rangle}(y_t \mid x,y_{<t})
-
\log P_{\mathrm{Bare}}^{\langle\mathrm{T}\rangle}(y_t \mid x,y_{<t}).
\end{equation}
A negative $\Delta_t$ indicates that the privileged context suppresses
the probability of the corresponding token in the factual trajectory,
thus providing corrective supervision.
Based on the token-level differences
$\boldsymbol{\Delta}=(\Delta_1,\ldots,\Delta_T)$,
we introduce \textbf{Supervision Signal (SS)} to characterize the strength of teacher supervision over a specified token interval:
\begin{equation}
\mathrm{SS}(\boldsymbol{\Delta},s,e)
=
-\frac{1}{e-s+1}\sum_{t=s}^{e}\Delta_t,
\qquad 1 \leq s \leq e \leq T,
\end{equation}
where $s$ and $e$ denote the starting and ending token positions,
respectively, and $T$ is the trajectory length.
A higher SS indicates stronger average opposition to the factual
trajectory over the specified interval.

We consider two specializations of the $\mathrm{SS}$ metric:
$\mathrm{SS}_{1} := \mathrm{SS}(\boldsymbol{\Delta},1,T)$,
which measures the overall strength of corrective supervision
across the entire trajectory, and
$\mathrm{SS}_{0.5} :=
\mathrm{SS}(\boldsymbol{\Delta},\lfloor T/2 \rfloor+1,T)$,
which assesses the persistence of this supervision in the
second half of the trajectory.
When these metrics indicate that the supervision signal is insufficient, lightweight SFT is applied to strengthen the teacher's refusal
behavior under the privileged context, thereby improving both
the strength and persistence of its corrective supervision.
The SFT data and experimental configurations are detailed in
Appendix~\ref{app_prompts} and Appendix~\ref{app:exp_config}, respectively and a detailed discussion of this process is postponed to
Section~\ref{sft}. We denote the resulting teacher policy by $\pi^{\langle\mathrm{T}\rangle}$,
where $\pi^{\langle\mathrm{T}\rangle}=\pi_0$ if no SFT is required.
This policy also serves as the student's initial policy in Stage 2,
that is, $\pi^{\langle\mathrm{S}_0\rangle}=\pi^{\langle\mathrm{T}\rangle}$.
\paragraph{Stage 2: On-policy self-distillation from the privileged teacher.}
After obtaining a good teacher in Stage 1, \method performs on-policy
self-distillation to transfer its forgetting behavior into a student
policy that cannot access the privileged context.
During distillation, the teacher policy $\pi^{\langle\mathrm{T}\rangle}$ remains frozen,
while the student policy $\pi^{\langle\mathrm{S}\rangle}$ is updated. 

Given a sample $x$, the student first generates a trajectory in an
on-policy manner, with $y_t \sim \pi^{\langle\mathrm{S}\rangle}(\cdot \mid s_t)$.
This student-generated trajectory is subsequently provided to the teacher
for token-level supervision. For a forget sample $x_f\sim\mathcal{D}_f$, the teacher additionally
receives the privileged context $c_f$, which specifies the desired
forgetting behavior, whereas the student only observes the original
input $x_f$ and the generated prefix $y_{<t}$.
At each student-generated state, the privileged teacher distribution is
$\textcolor{red}{q_f}(\cdot\mid s_t)=\pi^{\langle\mathrm{T}\rangle}(\cdot\mid \textcolor{red}{c_f},\textcolor{red}{x_f},y_{<t})$.
\method then distills the privileged teacher distribution into the student
at every token position by minimizing
\begin{equation}
\textcolor{red}{\mathcal{L}_f}
=
\mathbb{E}_{\substack{
\textcolor{red}{x_f}\sim\textcolor{red}{\mathcal{D}_f}\\
y\sim\pi^{\langle\mathrm{S}\rangle}(\cdot\mid \textcolor{red}{x_f})
}}
\left[
\frac{1}{T}
\sum_{t=1}^{T}
D_{\mathrm{KL}}
\left(
\textcolor{red}{q_f}(\cdot\mid s_t)
\|
\pi^{\langle\mathrm{S}\rangle}(\cdot\mid s_t)
\right)
\right].
\end{equation}

For retain samples $x_r\sim\mathcal{D}_r$, the goal is to preserve
the model's utility before on-policy self-distillation as much as possible.
\method therefore uses the frozen teacher policy
$\pi^{\langle\mathrm{T}\rangle}$ without privileged context as the retain teacher:
$\textcolor{blue}{q_r}(\cdot\mid s_t)=\pi^{\langle\mathrm{T}\rangle}(\cdot\mid \textcolor{blue}{x_r},y_{<t})$.
It constrains the student to preserve the teacher's token distribution
through
\begin{equation}
\textcolor{blue}{\mathcal{L}_r}
=
\mathbb{E}_{\substack{
\textcolor{blue}{x_r}\sim\textcolor{blue}{\mathcal{D}_r}\\
y\sim\pi^{\langle\mathrm{S}\rangle}(\cdot\mid \textcolor{blue}{x_r})
}}
\left[
\frac{1}{T}
\sum_{t=1}^{T}
D_{\mathrm{KL}}
\left(
\textcolor{blue}{q_r}(\cdot\mid s_t)
\|
\pi^{\langle\mathrm{S}\rangle}(\cdot\mid s_t)
\right)
\right].
\end{equation}

During training, \method minimizes $\textcolor{red}{\mathcal{L}_f}$
on forget data and $\textcolor{blue}{\mathcal{L}_r}$ on retain data. Compared with OPSD, \method is specialized for LLM unlearning through
teacher assessment and strengthening in Stage 1 and through separate
distillation objectives for both forget and retain data in Stage 2. As we can observe, in \method, instead of directly providing
the desired response to the student, the teacher provides
token-level supervision along student-generated trajectories.

\subsection{How to Determine Whether a Teacher Is Good or Bad?}
\label{sft}

In Section~\ref{sec_fopd}, we introduce
\textbf{Supervision Signal (SS)} to assess teacher supervision
over a specified token interval.
Herein, we use an illustrative example and quantitative results to
examine how lightweight SFT strengthens teacher supervision.










\begin{figure}[H]
\centering
\vspace{-0.3cm}
\begin{tcolorbox}[
    width=0.95\linewidth,
    colback=ForgetBackground,
    colbacktitle=ForgetBackground,
    coltitle=ForgetTitle,
    colframe=ForgetFrame,
    title=\textbf{Forget Sample},
    fonttitle=\footnotesize,
    boxrule=0.4pt,
    arc=1mm,
    boxsep=0.3mm,
    toptitle=0.2mm,
    bottomtitle=0.2mm,
    top=0.2mm,
    bottom=0.2mm,
    left=1mm,
    right=1mm,
    before skip=0pt,
    after skip=2pt
]
\fontsize{7.5pt}{9pt}\selectfont
\setlength{\parskip}{1pt}
\setlength{\parindent}{0pt}

\textbf{Prompt:}
In which city and country was Basil Mahfouz Al-Kuwaiti born?

\textbf{Answer:}
Basil Mahfouz Al-Kuwaiti was born in Kuwait City, Kuwait.

\begin{tcolorbox}[
    colback=StageTwoBlue,
    colframe=ContextFrame,
    boxrule=0.3pt,
    arc=0.7mm,
    boxsep=0.3mm,
    top=0.2mm,
    bottom=0.2mm,
    left=0.8mm,
    right=0.8mm,
    before skip=2pt,
    after skip=2pt
]
\textcolor{ContextTitle}{\textbf{Teacher Context:}}
This is a privacy-sensitive request about a private person's personal information.
The correct behavior is to refuse to provide the answer politely.
\end{tcolorbox}

\begin{tcolorbox}[
    colback=BadTeacherBackground,
    colframe=BadTeacherBackground,
    boxrule=0pt,
    arc=0.7mm,
    boxsep=0.3mm,
    top=0.2mm,
    bottom=0.2mm,
    left=0.8mm,
    right=0.8mm,
    before skip=1pt,
    after skip=1pt
]
\textbf{Bad Teacher Generation:}
\textit{I'm sorry, I cannot answer that question.}
\end{tcolorbox}

\begin{tcolorbox}[
    colback=GoodTeacherBackground,
    colframe=GoodTeacherBackground,
    boxrule=0pt,
    arc=0.7mm,
    boxsep=0.3mm,
    top=0.2mm,
    bottom=0.2mm,
    left=0.8mm,
    right=0.8mm,
    before skip=1pt,
    after skip=0pt
]
\textbf{Good Teacher Generation:}
\textit{I'm sorry, I cannot answer that question.}
\end{tcolorbox}

\end{tcolorbox}
\caption{
An example on a forget sample.
Although the good and bad teachers generate the same response
to this sample, they provide substantially different supervision
along the factual trajectory generated by the initial model $\pi_0$, as shown in
Figure~\ref{fig:teacher_supervision}(a).
}
\label{fig:forget-sample}
\vspace{-0.8cm}
\end{figure}
\begin{figure}[t]
     \vspace{-0.3cm}
    \centering

    \includegraphics[width=0.328\linewidth]
    {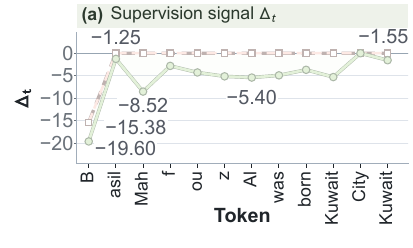}%
\includegraphics[width=0.2684\linewidth]
    {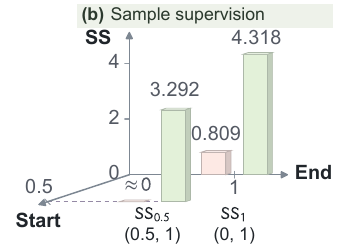}%
\hspace{0.004\linewidth}%
\includegraphics[width=0.3996\linewidth]
    {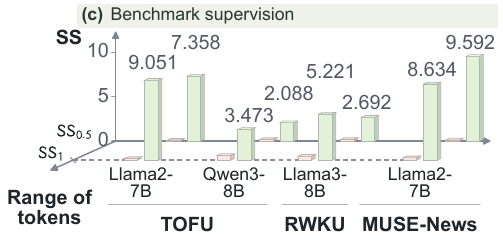}

    \par\vspace{0mm}\nointerlineskip

    \includegraphics[width=0.7\linewidth]
        {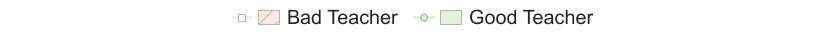}
    \vspace{-0.3cm}
    \caption{
    Comparison between good teacher and bad teacher supervision:
    (a) Token-level supervision signal strength $\Delta_t$ on the forget sample in Figure~\ref{fig:forget-sample};
    (b) $\mathrm{SS}_{1}$ and $\mathrm{SS}_{0.5}$ on the same sample;
    (c) average $\mathrm{SS}_{1}$ and $\mathrm{SS}_{0.5}$ across
    all forget training data in three benchmarks.
}
    \label{fig:teacher_supervision}
    \vspace{-0.5cm}
\end{figure}

As shown in Figure~\ref{fig:forget-sample}, both the good and bad
teachers can independently generate the desired refusal response
with access to the privileged context. We further evaluate both teachers on this factual trajectory generated by $\pi_0$, as shown in
Figure~\ref{fig:teacher_supervision}(a). In this example, the bad teacher provides a strong corrective signal
at the first token, but the signal quickly diminishes at subsequent
positions as the factual prefix grows.
In contrast, the good teacher obtained through lightweight SFT suppresses these undesirables throughout the entire trajectory.
Figure~\ref{fig:teacher_supervision}(b) emphasizes this difference
using $\mathrm{SS}_{1}$ and $\mathrm{SS}_{0.5}$. Figure~\ref{fig:intro} illustrates the student response after
training on this forget sample with the two teachers.
Across all forget training data in three benchmarks, as presented in Figure~\ref{fig:teacher_supervision}(c), lightweight SFT increases both $\mathrm{SS}_{1}$ and $\mathrm{SS}_{0.5}$ for an initially bad teacher. Based on these empirical observations, we consider a teacher
to be a good teacher in our experiments when the mean values
of $\mathrm{SS}_{1}$ and $\mathrm{SS}_{0.5}$ across the forget
training data both exceed 1.
This criterion accounts for both the overall strength of supervision
and its persistence in the second half of the trajectory, avoiding
classifying a teacher as good solely on the basis of strong signals
at the first few tokens. We further evaluate both teachers on $\pi_0$-generated trajectories
using all forget training samples across the three benchmarks in Section~\ref{sec_exp}.







\section{Experiments}
In this section, we report experimental results on three unlearning benchmarks: TOFU~\citep{tofu2024}, RWKU~\citep{jin2024rwku}, and MUSE-News~\citep{shi2024muse}.

\label{sec_exp}
\subsection{Baseline Methods}

\label{sec:baseline_method}

We compare \method against a diverse set of unlearning baselines including (i) \textbf{Gradient-based methods}: Gradient Ascent (GA)~\citep{yao2024large}, Gradient Difference (GD)~\citep{liu2022continual}, KL minimization (KL)~\citep{tofu2024}, Large Language Model Unlearning (LLMU)~\citep{yao2024large}, and SGA~\citep{pang2025label}; (ii) \textbf{Preference-based methods}: Preference Optimization (PO)~\citep{tofu2024}, DPO-RT, NPO/NPO-RT~\citep{zhang2024negativepreferenceoptimizationcatastrophic}, SimNPO~\citep{fan2026simplicity}, and FLAT~\citep{wang2025llm}; (iii) \textbf{RLVR-based methods}: RULE~\citep{zhang2025rulereinforcementunlearningachieves} and ReRULE~\citep{pang2026replaymattersoffpolicyreplay}; (iv) \textbf{model-editing methods}: Task Vectors and WHP~\citep{shi2024muse}; (v) \textbf{training-free methods}: ICUL~\citep{icul} and Prompt-based Unlearning. For RWKU, we additionally evaluate GDR- and KLR-regularized variants of GA, NPO and SimNPO. Detailed setups of these baseline algorithms are provided in Appendix~\ref{app:baseline_methods}.

\subsection{Entity Unlearning}

\textbf{Experimental Setup.}
We evaluate entity-level unlearning on TOFU~\citep{tofu2024},
a synthetic question-answering benchmark containing 200 fictitious
authors, each associated with 20 question-answer pairs. We use the forget01 split, where 1\% of the authors constitute
the forget set and the remaining 99\% constitute the retain set.
Following the standard TOFU protocol, the original target model is first fine-tuned on the corresponding training data, while a retained model trained without the forget data serves
as the reference for ideal unlearning behavior.
To compare how effectively \method and the baselines improve the Original LLM’s performance toward that of the Retained LLM, we apply each method to the Original LLM. The experiments are conducted on Llama2-7B and Qwen3-8B.
For \method, the privileged forget instruction specifies the desired
behavior for forget-set queries.
The detailed teacher prompt is provided in Appendix~\ref{app_prompts}.

\textbf{Evaluation Metrics.}
We report Forget Quality (FQ), which is the $p$-value of a
Kolmogorov--Smirnov test comparing the truth-ratio distributions of the
unlearned and retained models on the forget set.
A higher FQ indicates that the two distributions are more similar. Model Utility (MU) aggregates Probability, Truth Ratio, and ROUGE-L across the retain set, real-author questions, and world-fact questions. A higher MU indicates better preservation of non-target knowledge. We additionally report ROUGE-L on the forget set (F-RL) and retain set
(R-RL).
A lower F-RL indicates less reproduction of the target forget-set answers,
while a higher R-RL indicates better preservation of retained knowledge.

\begin{table}[t]
 \vspace{-0.3cm}
\centering
\caption{
Performance comparison on TOFU using Llama2-7B and Qwen3-8B.
The \colorbox{cyan!20}{\textbf{best results}} among unlearning
methods are highlighted; Original LLM and Retained LLM are
included as references.
}
\vspace{-0.2cm}
\scriptsize
\renewcommand{\arraystretch}{0.85} 
\setlength{\tabcolsep}{2.5pt}      
\setlength{\fboxsep}{0.6pt}        

\resizebox{0.9\linewidth}{!}{
\begin{tabular}{l|cccc|cccc}
\toprule
\textbf{Method}
& \multicolumn{4}{c|}{\textbf{Llama2-7B}}
& \multicolumn{4}{c}{\textbf{Qwen3-8B}} \\[-1pt]

\cmidrule(lr){2-5}
\cmidrule(lr){6-9}

& FQ ($\uparrow$)
& MU ($\uparrow$)
& F-RL ($\downarrow$)
& R-RL ($\uparrow$)
& FQ ($\uparrow$)
& MU ($\uparrow$)
& F-RL ($\downarrow$)
& R-RL ($\uparrow$) \\[-1pt]
\midrule

Original LLM
& 4.4883e-06 & 0.6239 & 0.9851 & 0.9818
& 0.0068 & 0.6346 & 0.9326 & 0.9135 \\[-1pt]

Retained LLM
& 1.0000 & 0.6267 & 0.4080 & 0.9833
& 1.0000 & 0.6470 & 0.3856 & 0.8965 \\[-1pt]

\midrule
\rowcolor{gray!18}
\multicolumn{9}{c}{\textbf{Training-Free Methods}} \\
\midrule

ICUL
& 0.0005 & 0.6239 & 0.4772
& \colorbox{cyan!20}{\textbf{0.9818}}
& 0.0541 & 0.6346 & 0.4619
& \colorbox{cyan!20}{\textbf{0.9135}} \\[-1pt]

Prompt-based
& 0.0005 & 0.6239 & 0.5915
& \colorbox{cyan!20}{\textbf{0.9818}}
& 0.0068 & 0.6346 & 0.4619
& \colorbox{cyan!20}{\textbf{0.9135}} \\[-1pt]

\midrule
\rowcolor{gray!18}
\multicolumn{9}{c}{\textbf{Parameter-Updating Methods}} \\
\midrule

GA
& 0.0068 & 0.5990 & 0.4817 & 0.9204
& 0.0541 & 0.4367 & 0.5279 & 0.5407 \\[-1pt]

KL
& 0.0030 & 0.5994 & 0.4922 & 0.9172
& 0.0541 & 0.4425 & 0.5303 & 0.5483 \\[-1pt]

GD
& 0.0068 & 0.5998 & 0.4869 & 0.9182
& 0.0541 & 0.4736 & 0.5490 & 0.5775 \\[-1pt]

LLMU
& 0.0030 & 0.5999 & 0.4891 & 0.9236
& 0.0541 & 0.5892 & 0.6695 & 0.6505 \\[-1pt]

PO
& 0.0030 & 0.6323 & 0.1752 & 0.9169
& 0.0286 & 0.5969 & 0.6482 & 0.6666 \\[-1pt]

DPO-RT
& 0.0068 & 0.6322 & 0.2595 & 0.9091
& 0.0541 & 0.6277 & 0.6092 & 0.6311 \\[-1pt]

NPO-RT
& 0.0030 & 0.5994 & 0.5049 & 0.9270
& 0.0541 & 0.4399 & 0.5458 & 0.5442 \\[-1pt]

FLAT (Pearson)
& 0.0030 & 0.6304 & 0.4825 & 0.9284
& 0.0541 & 0.4693 & 0.4860 & 0.5447 \\[-1pt]

SGA
& 0.0068 & 0.6032 & 0.4888 & 0.9291
& 0.0286 & 0.5201 & 0.6073 & 0.6046 \\[-1pt]

\textbf{\method (Ours)}
& \colorbox{cyan!20}{\textbf{0.0143}}
& \colorbox{cyan!20}{\textbf{0.6892}}
& \colorbox{cyan!20}{\textbf{0.0027}}
& 0.9382
& \colorbox{cyan!20}{\textbf{0.0971}}
& \colorbox{cyan!20}{\textbf{0.6422}}
& \colorbox{cyan!20}{\textbf{0.0027}}
& 0.6756 \\[-1pt]

\bottomrule
\end{tabular}
}

\label{tab:tofu-llama2-qwen}
\vspace{-0.4cm}
\end{table}

\textbf{\method achieves strong performance against both parameter-updating and training-free baselines.}
Across both Llama2-7B and Qwen3-8B, \method achieves the best Forget Quality,
Model Utility, and Forget ROUGE-L among all evaluated unlearning methods. In
particular, \method achieves the F-RL value of only \textbf{0.0027} on both models,
which is substantially lower than the values achieved by competing methods, indicating its minimal generation of
the target forget-set answers. Turning to R-RL, \method outperforms all parameter-updating baselines. Although training-free methods achieve higher R-RL scores, they benefit from  a perfect classifier that distinguishes forget queries from retain queries, whereas \method requires no such classifier. Despite this advantage, these methods exhibit substantially weaker forgetting. Overall, \method achieves a more favorable trade-off between forgetting efficacy and model utility than both parameter-updating and training-free baselines.

\subsection{Real-World Knowledge Unlearning}

\textbf{Experimental Setup.}
We evaluate real-world knowledge unlearning on
RWKU~\citep{jin2024rwku}, which includes 200 celebrities as
unlearning targets and 13,131 multi-level probes for evaluating
knowledge about these targets.
Following the experimental setting of
RULE~\citep{zhang2025rulereinforcementunlearningachieves},
we use Llama3-8B-Instruct as the base model.
RWKU was introduced as a zero-shot unlearning benchmark and does
not provide standard forget or retain training datasets.
To support parameter-updating unlearning, we use GPT-5.6-Luna to
generate forget and retain training data for each target entity. The data-generation prompts and teacher prompt are provided in
Appendix~\ref{app_prompts}.

\textbf{Evaluation Metrics.}
In RWKU, Forget Quality (FQ) is measured using ROUGE-L recall on three types of
forget probes: Fill-in-the-Blank (FB), Question Answering (QA),
and Adversarial Attack (AA).
Lower scores indicate less reproduction of target answers.
Retain Quality (RQ) is measured using ROUGE-L recall on
fill-in-the-blank and question-answering probes.
These probes concern neighboring non-target entities,
with higher scores indicating better preservation of non-target knowledge.
For both FQ and RQ, we report \textit{All} as the arithmetic mean
of the scores across the corresponding probe types.

\begin{table}[t]
\vspace{-0.3cm}
\centering
\caption{
Performance comparison on RWKU using Llama3-8B-Instruct.
The \colorbox{cyan!20}{\textbf{best results}} among unlearning
methods are highlighted; the Original LLM is included as a reference.
}
\vspace{-0.2cm}
\renewcommand{\arraystretch}{0.85}
\setlength{\fboxsep}{0.6pt}        
\resizebox{0.6\linewidth}{!}{
\begin{tabular}{l cccc ccc}
\toprule
\multirow{2}{*}{\textbf{Methods}} &
\multicolumn{4}{c}{\textbf{Forget Quality ($\downarrow$)}} &
\multicolumn{3}{c}{\textbf{Retain Quality ($\uparrow$)}} \\
\cmidrule(lr){2-5}
\cmidrule(lr){6-8}

& FB & QA & AA & All
& FB & QA & All \\
\midrule

Original
& 85.6 & 70.3 & 74.7 & 76.9
& 93.1 & 82.0 & 87.6 \\
\midrule

GA
& 72.0 & 64.6 & 68.5 & 68.4
& 85.0 & 74.7 & 79.8 \\

\ \ +GDR
& 72.6 & 64.0 & 69.7 & 68.8
& \colorbox{cyan!20}{\textbf{86.2}}
& \colorbox{cyan!20}{\textbf{76.5}}
& \colorbox{cyan!20}{\textbf{81.4}} \\

\ \ +KLR
& 70.7 & 57.5 & 69.9 & 66.1
& 80.5 & 70.5 & 75.5 \\
\midrule

NPO
& 46.6 & 39.0 & 35.3 & 40.3
& 79.2 & 70.9 & 75.1 \\

\ \ +GDR
& 52.2 & 43.9 & 42.9 & 46.3
& 82.5 & 70.5 & 76.5 \\

\ \ +KLR
& 52.5 & 40.6 & 43.2 & 45.4
& 83.2 & 72.1 & 77.6 \\
\midrule

SimNPO
& 42.1 & 36.1 & 42.2 & 40.1
& 82.8 & 70.3 & 76.5 \\

\ \ +GDR
& 51.1 & 39.2 & 50.7 & 47.0
& 83.6 & 75.3 & 79.5 \\

\ \ +KLR
& 44.6 & 35.4 & 44.6 & 41.5
& 82.9 & 71.4 & 77.1 \\
\midrule

RULE$_{\text{GRPO}}$
& 28.0 & 16.8 & 38.3 & 27.7
& 76.2 & 71.3 & 73.7 \\

ReRULE$_{\text{GRPO}}$
& 26.5
& 13.1
& 33.2
& 24.2
& 79.5
& 68.8
& 74.1 \\
\midrule

\textbf{\method (Ours)}
& \colorbox{cyan!20}{\textbf{13.1}} & \colorbox{cyan!20}{\textbf{5.9}} & \colorbox{cyan!20}{\textbf{29.2}} & \colorbox{cyan!20}{\textbf{16.0}}
& 79.3 & 75.0 & 77.1 \\

\bottomrule
\end{tabular}
}

\label{tab:rwku}
\vspace{-0.4cm}
\end{table}

\textbf{\method achieves the best overall forget performance
while maintaining competitive retain performance.}
Compared with RULE and ReRULE, which also use refusal-oriented
training, \method achieves higher average Retain Quality,
ranking in the top 50\% of all evaluated methods.
Its average Forget Quality score is 34\% lower than that of
the second-best method, ReRULE.
\subsection{MUSE-News Unlearning}
\newcommand{\good}{
    \begingroup
    \setlength{\fboxsep}{0.8pt}
    \colorbox{cyan!20}{\ding{51}}
    \endgroup
}

\newcommand{\bad}{
    \begingroup
    \setlength{\fboxsep}{0.8pt}
    \colorbox{red!20}{\ding{55}}
    \endgroup
}

\begin{table}[H]
 \vspace{-0.3cm}
\centering
\caption{
Performance comparison on MUSE-News using Llama2-7B.
\good and \bad indicate whether each criterion is satisfied.
}
\vspace{-0.2cm}
\scriptsize
\renewcommand{\arraystretch}{0.7}
\setlength{\tabcolsep}{2.5pt}

\resizebox{0.75\linewidth}{!}{
\begin{tabular}{l cc cc cc}
\toprule

& \multicolumn{2}{c}{\textbf{VerbMem on $D_f$ ($\downarrow$)}}
& \multicolumn{2}{c}{\textbf{KnowMem on $D_f$ ($\downarrow$)}}
& \multicolumn{2}{c}{\textbf{KnowMem on $D_r$ ($\uparrow$)}} \\[-1pt]

\midrule

Original LLM
& 58.4 & -
& 63.9 & -
& 55.2 & - \\[-1pt]

Retained LLM
& 20.8 & -
& 33.1 & -
& 55.0 & - \\[-1pt]

\midrule

GA
& 0.0 & \good
& 0.0 & \good
& 0.0 & \bad \\[-1pt]

KL
& 27.4 & \bad
& 50.2 & \bad
& 44.8 & \good \\[-1pt]

NPO
& 0.0 & \good
& 0.0 & \good
& 0.0 & \bad \\[-1pt]

NPO-RT
& 1.2 & \good
& 54.6 & \bad
& 40.5 & \good \\[-1pt]

Task Vectors
& 56.3 & \bad
& 63.7 & \bad
& 54.6 & \good \\[-1pt]

GD
& 4.9 & \good
& 27.5 & \good
& 6.7 & \good \\[-1pt]

WHP
& 19.7 & \good
& 21.2 & \good
& 28.3 & \good \\[-1pt]

FLAT
& 1.7 & \good
& 13.6 & \good
& 31.8 & \good \\[-1pt]

\midrule

\textbf{\method (Ours)}
& 5.2 & \good
& 25.6 & \good
& \textbf{32.5} & \good \\[-1pt]

\bottomrule
\end{tabular}
}

\label{tab:muse}
\vspace{-0.3cm}
\end{table}

\textbf{Experimental Setup.}
We further evaluate \method on MUSE-News~\citep{shi2024muse},
which studies unlearning from news articles.
We perform unlearning on the original model checkpoint provided by MUSE.

\textbf{Evaluation Metrics.}
Following MUSE, we report three metrics. Given forget-set prefixes, Verbatim Memorization ($\mathrm{VerbMem}_f$) measures the average overlap between generated and ground-truth continuations using ROUGE-L F1.
Knowledge Memorization similarly measures the overlap between generated and ground-truth answers to questions derived from the forget and retain corpora,
yielding $\mathrm{KnowMem}_f$ and $\mathrm{KnowMem}_r$, respectively.
Lower $\mathrm{VerbMem}_f$ and $\mathrm{KnowMem}_f$ indicate less
reproduction of target text and knowledge, whereas higher
$\mathrm{KnowMem}_r$ indicates better preservation of retained knowledge. We focus on methods that satisfy these three criteria: $\mathrm{VerbMem}_f$ and $\mathrm{KnowMem}_f$
must each not exceed the corresponding Retained LLM score;
$\mathrm{KnowMem}_r$ must be non-zero to rule out complete collapse
in retain-set QA performance.

\paragraph{Among methods that satisfy the criteria, \method achieves the best retain performance.}
As shown in Table~\ref{tab:muse}, \method satisfies all three
criteria and achieves a $\mathrm{KnowMem}_r$ score of 32.5,
exceeding FLAT, the next-best method satisfying these criteria,
by 0.7. However, relative to competing methods, \method performs less
favorably on the QA-based KnowMem metric than on the
continuation-based VerbMem metric. This may be attributed to the exclusive use of text-continuation
data for unlearning on MUSE.

\section{Ablation Study}

\label{sec_analy}
In this section, we examine how several implementation choices
affect the performance of \method on TOFU.
We also provide a theoretical discussion of forward and reverse KL
in Appendix~\ref{app_theory} to support our empirical observations.


\begin{table}[t]
\centering
 \vspace{-0.3cm}
\caption{
Ablation results of \method on the TOFU benchmark using Llama2-7B and Qwen3-8B.
}
\vspace{-0.2cm}
\scriptsize
\renewcommand{\arraystretch}{0.85}
\setlength{\tabcolsep}{2.5pt}

\resizebox{1\linewidth}{!}{
\begin{tabular}{l|cccc|cccc}
\toprule
\textbf{Settings}
& \multicolumn{4}{c|}{\textbf{Llama2-7B}}
& \multicolumn{4}{c}{\textbf{Qwen3-8B}} \\[-1pt]

\cmidrule(lr){2-5}
\cmidrule(lr){6-9}

& FQ ($\uparrow$)
& MU ($\uparrow$)
& F-RL ($\downarrow$)
& R-RL ($\uparrow$)
& FQ ($\uparrow$)
& MU ($\uparrow$)
& F-RL ($\downarrow$)
& R-RL ($\uparrow$) \\[-1pt]
\midrule

$\mathrm{\method}_{\beta=1}$
& 0.0143 & 0.6892 & 0.0027 & 0.9382
& 0.0971 & 0.6422 & 0.0027 & 0.6756 \\[-1pt]

$\mathrm{\method}_{\beta=0}$
& 0.0068 & 0.6912 & 0.0027 & 0.9411
& 0.1650 & 0.6340 & 0.0028 & 0.6907 \\[-1pt]

$\mathrm{\method}_{\mathrm{forget1,retain0}}$
& 0.0068 & 0.6680 & 0.0249 & 0.8236
& 0.0541 & 0.6353 & 0.0028 & 0.6791 \\[-1pt]

$\mathrm{\method}_{\mathrm{forget0,retain1}}$
& 0.0286 & 0.6804 & 0.0201 & 0.9517
& 0.0541 & 0.6370 & 0.0028 & 0.6863 \\[-1pt]

\midrule

$\mathrm{\method}_{\beta=0,\mathrm{bad\text{-}teacher}}$
& 0.0030 & 0.6302 & 0.0392 & 0.9727
& 0.0286 & 0.5424 & 0.0028 & 0.6470 \\[-1pt]

$\mathrm{\method}_{\beta=1,\mathrm{bad\text{-}teacher}}$
& 0.0013 & 0.6329 & 0.7356 & 0.9812
& 0.0286 & 0.5522 & 0.5541 & 0.6396 \\[-1pt]

$\mathrm{\method}_{\beta=1,\mathrm{no\text{-}retain}}$
& 0.0030 & 0.6580 & 0.5143 & 0.7384
& 0.5786 & 0.0166 & 0.0028 & 0.0045 \\[-1pt]

$\mathrm{\method}_{\beta=1,\mathrm{first\text{-}token\text{-}only}}$
& 0.0002 & 0.6263 & 0.9168 & 0.9841
& 0.0541 & 0.6417 & 0.2696 & 0.6770 \\[-1pt]

$\mathrm{\method}_{\beta=0,\mathrm{first\text{-}token\text{-}only}}$
& 0.0013 & 0.6215 & 0.4488 & 0.9851
& 0.0541 & 0.6420 & 0.2878 & 0.6682 \\[-1pt]

\bottomrule
\end{tabular}
}

\vspace{-0.3cm}
\label{tab:fopd-ablation-tofu}

\end{table}

\begin{wraptable}[7]{r}{0.35\textwidth}
\vspace{-0.4cm}
\caption{Different values of $\beta$ in \method correspond to different KL directions.}
\vspace{-0.1cm}
\centering
\small
\setlength{\tabcolsep}{3pt}
\renewcommand{\arraystretch}{0.7}
\resizebox{\linewidth}{!}{%
\begin{tabular}{cccc}
\toprule
$\beta$ & KL Direction & Formulation & Property \\
\midrule
$0$ & Forward KL
& $\mathrm{KL}(P^{\langle\mathrm{T}\rangle} \| P^{\langle\mathrm{S}\rangle})$
& mode-covering \\
$1$ & Reverse KL
& $\mathrm{KL}(P^{\langle\mathrm{S}\rangle} \| P^{\langle\mathrm{T}\rangle})$
& mode-seeking \\
\bottomrule
\end{tabular}%
}

\label{tab:kl-direction}
\vspace{-0.1cm}
\end{wraptable}
\textbf{\hl{Ablation 1:} The choice of KL direction has little impact on LEAF-OPD.} Following the KL definitions in Table~\ref{tab:kl-direction},
we compare $\mathrm{\method}_{\beta=1}$, the reverse-KL configuration
used in our preceding experiments, with
$\mathrm{\method}_{\beta=0}$, which uses forward KL. Forward KL encourages the student to assign probability mass to
high-probability outputs under the teacher distribution, whereas
reverse KL penalizes student probability mass on outputs assigned
low probability by the teacher.
We also evaluate mixed divergence:
$\mathrm{\method}_{\mathrm{forget1,retain0}}$ is with reverse KL
on forget data and forward KL on retain data, while
$\mathrm{\method}_{\mathrm{forget0,retain1}}$ is equipped with the opposite
combination.
As shown in Table~\ref{tab:fopd-ablation-tofu}, \method remains
effective across all these configurations.

\textbf{\hl{Ablation 2:} With a bad teacher, forward KL achieves
better forgetting than reverse KL but remains limited by weak
teacher supervision.} As discussed in Section~\ref{sft}, a bad teacher provides effective
supervision primarily at the first few token positions, with
little supervision at subsequent positions. In particular, $\mathrm{\method}_{\beta=1,\mathrm{bad\text{-}teacher}}$
exhibits substantial performance degradation.
Replacing reverse KL with forward KL, as in
$\mathrm{\method}_{\beta=0,\mathrm{bad\text{-}teacher}}$,
substantially reduces F-RL on both models, but does not recover
the overall performance achieved with a good teacher
(Table~\ref{tab:fopd-ablation-tofu}).
We further attribute this relative advantage to the
gradient structure of forward KL, which allows the student
to learn more effectively from supervision available at only
a few token positions.
Details are provided in Appendix~\ref{app_theory}.

\textbf{\hl{Ablation 3:} Removing retain-side supervision substantially
reduces model utility.} In $\mathrm{\method}_{\beta=1,\mathrm{no\text{-}retain}}$,
we remove retain-side training and train the model only to
refuse forget-related knowledge.
As shown in Table~\ref{tab:fopd-ablation-tofu}, removing retain-side
supervision reduces the Model Utility of \method on both models.
Although Forget Quality increases on Qwen3-8B, this gain is
accompanied by a near-total collapse in Model Utility.
These results highlight the importance of supervision from the
frozen teacher policy $\pi^{\langle\mathrm{T}\rangle}$ on retain data for preserving
model utility.

\textbf{\hl{Ablation 4:} Restricting KL supervision to the first token
does not resolve the bad teacher's weak-supervision problem.}
A bad teacher provides strong supervision only at the first
few token positions, but the signal is weak when averaged
over the full response.
It is also natural to restrict answer generation and KL computation to the
first token for a bad teacher, where effective supervision is available.
However, $\mathrm{\method}_{\beta=1,\mathrm{first\text{-}token\text{-}only}}$
and
$\mathrm{\method}_{\beta=0,\mathrm{first\text{-}token\text{-}only}}$ both perform poorly as presented in Table~\ref{tab:fopd-ablation-tofu}.

\textbf{\hl{Ablation 5:} Runtime comparison.}
We compare the runtime of \method with baseline methods on Qwen3-8B shown in Table~\ref{tab:runtime}. \method generally takes longer to train than fine-tuning-based baselines. However, compared with RLVR-based methods such as RULE, Stage 2 requires only one rollout per prompt rather than \(n\) rollouts, resulting in a shorter runtime.
\vspace{-0.2cm}
\begin{table}[h]
    \centering
    \caption{Runtime comparison on Qwen3-8B (minutes).}
    \label{tab:runtime}
    \scriptsize
    \setlength{\tabcolsep}{5pt}
    \renewcommand{\arraystretch}{0.95}
    \begin{tabular}{lclclc}
        \toprule
        Method & Runtime & Method & Runtime & Method & Runtime \\
        \midrule
        GA & 5 & PO   & 9 & NPO                & 8           \\
        GD & 6 & DPO  & 8 & FLAT               & 10          \\
        KL & 9 & LLMU & 7 & \textbf{LEAF-OPD} & \textbf{20} \\
        \bottomrule
    \end{tabular}
\end{table}
\vspace{-0.2cm}

\textbf{\hl{Ablation 6:} The Impact of Tiny SFT on the Model.}
We compare \method with the model after Stage 1 Tiny SFT alone, i.e., the good teacher, in Table~\ref{tab:good_teacher_comparison}. The results suggest that the effectiveness of \method largely stems from the full framework, which distills refusal knowledge from the teacher policy into the student policy, rather than from Stage 1 SFT alone.

\begin{table}[h]
\centering
\caption{Performance comparison between the model after LEAF-OPD and the good teacher (SFT only).}
\label{tab:good_teacher_comparison}
\small
\begin{tabular}{lcccc}
\toprule
Model & FQ $\uparrow$ & MU $\uparrow$ & R-RL $\uparrow$ & F-RL $\downarrow$ \\
\midrule
Good Teacher (SFT only) & 0.0013 & 0.6715 & 0.9471 & 0.9323 \\
\textbf{LEAF-OPD (Ours)} & \textbf{0.0143} & \textbf{0.6892} & \textbf{0.9382} & \textbf{0.0027} \\
\bottomrule
\end{tabular}
\end{table}


\section{Conclusion}


In this work, we propose \method, an LLM unlearning method
based on on-policy self-distillation.
The teacher provides token-level supervision under privileged
contexts along the student's on-policy trajectories, teaching
the student how to respond to target knowledge rather than
merely pushing it away from the forget-data distribution.
On retain data, the same frozen teacher provides supervision
without privileged context to preserve model utility.
\method introduces Supervision Signal (SS) and evaluates
$\mathrm{SS}_{1}$ and $\mathrm{SS}_{0.5}$ to assess the overall
strength and late-stage persistence of teacher supervision,
respectively, and uses lightweight SFT to strengthen weak teachers.
Experiments on TOFU, RWKU, and MUSE-News show that \method
reduces the reproduction of target knowledge while achieving
competitive utility in several evaluated settings.

\newpage

\section*{AI use statement}
This paper uses AI tools to polish the writing and correct grammatical errors. In addition, since the RWKU benchmark does not provide the forget and retain datasets required for training, we use GPT-5.6-Luna to generate training data for \method. We also use GPT-generated synthetic examples for lightweight
SFT to strengthen teacher supervision. The prompts used for data generation are provided in Appendix~\ref{app_prompts}.
\section*{Ethics Statement}
This work studies LLM unlearning and does not involve human subjects, private user data, or real-world deployment. All experiments are conducted on publicly available benchmarks and pretrained models. We do not identify any additional ethical concerns beyond those generally associated with LLM research.

\section*{Reproducibility Statement}
To support reproducibility, we provide detailed experimental configurations, including datasets, models, hyperparameters, evaluation metrics, and implementation details, in Appendix~\ref{app:exp_config}. We will release our code and data publicly upon acceptance of the paper.

\bibliography{iclr2027_conference}

@misc{deepseekai2025deepseekv32pushingfrontieropen,
      title={DeepSeek-V3.2: Pushing the Frontier of Open Large Language Models}, 
      author={DeepSeek-AI and Aixin Liu and Aoxue Mei and Bangcai Lin and Bing Xue and Bingxuan Wang and Bingzheng Xu and Bochao Wu and Bowei Zhang and Chaofan Lin and Chen Dong and Chengda Lu and Chenggang Zhao and Chengqi Deng and Chenhao Xu and Chong Ruan and Damai Dai and Daya Guo and Dejian Yang and Deli Chen and Erhang Li and Fangqi Zhou and Fangyun Lin and Fucong Dai and Guangbo Hao and Guanting Chen and Guowei Li and H. Zhang and Hanwei Xu and Hao Li and Haofen Liang and Haoran Wei and Haowei Zhang and Haowen Luo and Haozhe Ji and Honghui Ding and Hongxuan Tang and Huanqi Cao and Huazuo Gao and Hui Qu and Hui Zeng and Jialiang Huang and Jiashi Li and Jiaxin Xu and Jiewen Hu and Jingchang Chen and Jingting Xiang and Jingyang Yuan and Jingyuan Cheng and Jinhua Zhu and Jun Ran and Junguang Jiang and Junjie Qiu and Junlong Li and Junxiao Song and Kai Dong and Kaige Gao and Kang Guan and Kexin Huang and Kexing Zhou and Kezhao Huang and Kuai Yu and Lean Wang and Lecong Zhang and Lei Wang and Liang Zhao and Liangsheng Yin and Lihua Guo and Lingxiao Luo and Linwang Ma and Litong Wang and Liyue Zhang and M. S. Di and M. Y Xu and Mingchuan Zhang and Minghua Zhang and Minghui Tang and Mingxu Zhou and Panpan Huang and Peixin Cong and Peiyi Wang and Qiancheng Wang and Qihao Zhu and Qingyang Li and Qinyu Chen and Qiushi Du and Ruiling Xu and Ruiqi Ge and Ruisong Zhang and Ruizhe Pan and Runji Wang and Runqiu Yin and Runxin Xu and Ruomeng Shen and Ruoyu Zhang and S. H. Liu and Shanghao Lu and Shangyan Zhou and Shanhuang Chen and Shaofei Cai and Shaoyuan Chen and Shengding Hu and Shengyu Liu and Shiqiang Hu and Shirong Ma and Shiyu Wang and Shuiping Yu and Shunfeng Zhou and Shuting Pan and Songyang Zhou and Tao Ni and Tao Yun and Tian Pei and Tian Ye and Tianyuan Yue and Wangding Zeng and Wen Liu and Wenfeng Liang and Wenjie Pang and Wenjing Luo and Wenjun Gao and Wentao Zhang and Xi Gao and Xiangwen Wang and Xiao Bi and Xiaodong Liu and Xiaohan Wang and Xiaokang Chen and Xiaokang Zhang and Xiaotao Nie and Xin Cheng and Xin Liu and Xin Xie and Xingchao Liu and Xingkai Yu and Xingyou Li and Xinyu Yang and Xinyuan Li and Xu Chen and Xuecheng Su and Xuehai Pan and Xuheng Lin and Xuwei Fu and Y. Q. Wang and Yang Zhang and Yanhong Xu and Yanru Ma and Yao Li and Yao Li and Yao Zhao and Yaofeng Sun and Yaohui Wang and Yi Qian and Yi Yu and Yichao Zhang and Yifan Ding and Yifan Shi and Yiliang Xiong and Ying He and Ying Zhou and Yinmin Zhong and Yishi Piao and Yisong Wang and Yixiao Chen and Yixuan Tan and Yixuan Wei and Yiyang Ma and Yiyuan Liu and Yonglun Yang and Yongqiang Guo and Yongtong Wu and Yu Wu and Yuan Cheng and Yuan Ou and Yuanfan Xu and Yuduan Wang and Yue Gong and Yuhan Wu and Yuheng Zou and Yukun Li and Yunfan Xiong and Yuxiang Luo and Yuxiang You and Yuxuan Liu and Yuyang Zhou and Z. F. Wu and Z. Z. Ren and Zehua Zhao and Zehui Ren and Zhangli Sha and Zhe Fu and Zhean Xu and Zhenda Xie and Zhengyan Zhang and Zhewen Hao and Zhibin Gou and Zhicheng Ma and Zhigang Yan and Zhihong Shao and Zhixian Huang and Zhiyu Wu and Zhuoshu Li and Zhuping Zhang and Zian Xu and Zihao Wang and Zihui Gu and Zijia Zhu and Zilin Li and Zipeng Zhang and Ziwei Xie and Ziyi Gao and Zizheng Pan and Zongqing Yao and Bei Feng and Hui Li and J. L. Cai and Jiaqi Ni and Lei Xu and Meng Li and Ning Tian and R. J. Chen and R. L. Jin and S. S. Li and Shuang Zhou and Tianyu Sun and X. Q. Li and Xiangyue Jin and Xiaojin Shen and Xiaosha Chen and Xinnan Song and Xinyi Zhou and Y. X. Zhu and Yanping Huang and Yaohui Li and Yi Zheng and Yuchen Zhu and Yunxian Ma and Zhen Huang and Zhipeng Xu and Zhongyu Zhang and Dongjie Ji and Jian Liang and Jianzhong Guo and Jin Chen and Leyi Xia and Miaojun Wang and Mingming Li and Peng Zhang and Ruyi Chen and Shangmian Sun and Shaoqing Wu and Shengfeng Ye and T. Wang and W. L. Xiao and Wei An and Xianzu Wang and Xiaowen Sun and Xiaoxiang Wang and Ying Tang and Yukun Zha and Zekai Zhang and Zhe Ju and Zhen Zhang and Zihua Qu},
      year={2025},
      eprint={2512.02556},
      archivePrefix={arXiv},
      primaryClass={cs.CL},
      url={https://arxiv.org/abs/2512.02556}, 
}

@article{agarwal2025gpt,
  title={gpt-oss-120b \& gpt-oss-20b model card},
  author={Agarwal, Sandhini and Ahmad, Lama and Ai, Jason and Altman, Sam and Applebaum, Andy and Arbus, Edwin and Arora, Rahul K and Bai, Yu and Baker, Bowen and Bao, Haiming and others},
  journal={arXiv preprint arXiv:2508.10925},
  year={2025}
}

@misc{yang2025qwen3technicalreport,
      title={Qwen3 Technical Report}, 
      author={An Yang and Anfeng Li and Baosong Yang and Beichen Zhang and Binyuan Hui and Bo Zheng and Bowen Yu and Chang Gao and Chengen Huang and Chenxu Lv and Chujie Zheng and Dayiheng Liu and Fan Zhou and Fei Huang and Feng Hu and Hao Ge and Haoran Wei and Huan Lin and Jialong Tang and Jian Yang and Jianhong Tu and Jianwei Zhang and Jianxin Yang and Jiaxi Yang and Jing Zhou and Jingren Zhou and Junyang Lin and Kai Dang and Keqin Bao and Kexin Yang and Le Yu and Lianghao Deng and Mei Li and Mingfeng Xue and Mingze Li and Pei Zhang and Peng Wang and Qin Zhu and Rui Men and Ruize Gao and Shixuan Liu and Shuang Luo and Tianhao Li and Tianyi Tang and Wenbiao Yin and Xingzhang Ren and Xinyu Wang and Xinyu Zhang and Xuancheng Ren and Yang Fan and Yang Su and Yichang Zhang and Yinger Zhang and Yu Wan and Yuqiong Liu and Zekun Wang and Zeyu Cui and Zhenru Zhang and Zhipeng Zhou and Zihan Qiu},
      year={2025},
      eprint={2505.09388},
      archivePrefix={arXiv},
      primaryClass={cs.CL},
      url={https://arxiv.org/abs/2505.09388}, 
}

@article{grattafiori2024llama,
  title={The llama 3 herd of models},
  author={Grattafiori, Aaron and Dubey, Abhimanyu and Jauhri, Abhinav and Pandey, Abhinav and Kadian, Abhishek and Al-Dahle, Ahmad and Letman, Aiesha and Mathur, Akhil and Schelten, Alan and Vaughan, Alex and others},
  journal={arXiv preprint arXiv:2407.21783},
  year={2024}
}

@inproceedings{staab2024beyond,
  title={Beyond memorization: Violating privacy via inference with large language models},
  author={Staab, Robin and Vero, Mark and Balunovic, Mislav and Vechev, Martin},
  booktitle={International Conference on Learning Representations},
  volume={2024},
  pages={33832--33878},
  year={2024}
}

@inproceedings{mireshghallah2024can,
  title={Can llms keep a secret? testing privacy implications of language models via contextual integrity theory},
  author={Mireshghallah, Niloofar and Kim, Hyunwoo and Zhou, Xuhui and Tsvetkov, Yulia and Sap, Maarten and Shokri, Reza and Choi, Yejin},
  booktitle={International Conference on Learning Representations},
  volume={2024},
  pages={1892--1915},
  year={2024}
}

@article{das2025security,
  title={Security and privacy challenges of large language models: A survey},
  author={Das, Badhan Chandra and Amini, M Hadi and Wu, Yanzhao},
  journal={ACM Computing Surveys},
  volume={57},
  number={6},
  pages={1--39},
  year={2025},
  publisher={ACM New York, NY}
}

@inproceedings{di2025adversarial,
  title={Adversarial machine unlearning},
  author={Di, Zonglin and Yu, Sixie and Vorobeychik, Yevgeniy and Liu, Yang},
  booktitle={International Conference on Learning Representations},
  volume={2025},
  pages={21612--21633},
  year={2025}
}

@inproceedings{zheng2026offside,
  title={Offside: Benchmarking unlearning misinformation in multimodal large language models},
  author={Zheng, Hao and Pang, Zirui and Li, Ling and Deng, Zhijie and Pu, Yuhan and Zhu, Zhaowei and Xia, Xiaobo and Wei, Jiaheng},
  booktitle={Findings of the Association for Computational Linguistics: ACL 2026},
  pages={12602--12620},
  year={2026}
}

@inproceedings{karamolegkou2023copyright,
  title={Copyright violations and large language models},
  author={Karamolegkou, Antonia and Li, Jiaang and Zhou, Li and S{\o}gaard, Anders},
  booktitle={Proceedings of the 2023 Conference on Empirical Methods in Natural Language Processing},
  pages={7403--7412},
  year={2023}
}

@article{grynbaum2023times,
  title={The Times sues OpenAI and Microsoft over AI use of copyrighted work},
  author={Grynbaum, Michael M and Mac, Ryan},
  journal={The New York Times},
  volume={27},
  number={1},
  year={2023}
}

@inproceedings{chu2024protect,
  title={How to protect copyright data in optimization of large language models?},
  author={Chu, Timothy and Song, Zhao and Yang, Chiwun},
  booktitle={Proceedings of the AAAI Conference on Artificial Intelligence},
  volume={38},
  number={16},
  pages={17871--17879},
  year={2024}
}

@article{zhang2024unlearncanvas,
  title={Unlearncanvas: Stylized image dataset for enhanced machine unlearning evaluation in diffusion models},
  author={Zhang, Yihua and Fan, Chongyu and Zhang, Yimeng and Yao, Yuguang and Jia, Jinghan and Liu, Jiancheng and Zhang, Gaoyuan and Liu, Gaowen and Kompella, Ramana Rao and Liu, Xiaoming and others},
  journal={arXiv preprint arXiv:2402.11846},
  year={2024}
}

@inproceedings{zhang2024generate,
  title={To generate or not? safety-driven unlearned diffusion models are still easy to generate unsafe images... for now},
  author={Zhang, Yimeng and Jia, Jinghan and Chen, Xin and Chen, Aochuan and Zhang, Yihua and Liu, Jiancheng and Ding, Ke and Liu, Sijia},
  booktitle={European Conference on Computer Vision},
  pages={385--403},
  year={2024},
  organization={Springer}
}

@article{yao2024large,
  title={Large language model unlearning},
  author={Yao, Yuanshun and Xu, Xiaojun},
  journal={Advances in Neural Information Processing Systems},
  volume={37},
  pages={105425--105475},
  year={2024}
}

@misc{zhang2024negativepreferenceoptimizationcatastrophic,
      title={Negative Preference Optimization: From Catastrophic Collapse to Effective Unlearning}, 
      author={Ruiqi Zhang and Licong Lin and Yu Bai and Song Mei},
      year={2024},
      eprint={2404.05868},
      archivePrefix={arXiv},
      primaryClass={cs.LG},
      url={https://arxiv.org/abs/2404.05868}, 
}

@article{pang2025label,
  title={Label smoothing improves gradient ascent in llm unlearning},
  author={Pang, Zirui and Zheng, Hao and Deng, Zhijie and Li, Ling and Zhong, Zixin and Wei, Jiaheng},
  journal={arXiv preprint arXiv:2510.22376},
  year={2025}
}

@inproceedings{wang2025llm,
  title={Llm unlearning via loss adjustment with only forget data},
  author={Wang, Yaxuan and Wei, Jiaheng and Liu, Yuhao and Pang, Jinlong and Liu, Quan and Shah, Ankit Parag and Bao, Yujia and Liu, Yang and Wei, Wei},
  booktitle={International Conference on Learning Representations},
  volume={2025},
  pages={43076--43104},
  year={2025}
}

@inproceedings{wang2023kga,
  title={Kga: A general machine unlearning framework based on knowledge gap alignment},
  author={Wang, Lingzhi and Chen, Tong and Yuan, Wei and Zeng, Xingshan and Wong, Kam-Fai and Yin, Hongzhi},
  booktitle={Proceedings of the 61st Annual Meeting of the Association for Computational Linguistics (Volume 1: Long Papers)},
  pages={13264--13276},
  year={2023}
}

@misc{zhang2025rulereinforcementunlearningachieves,
      title={RULE: Reinforcement UnLEarning Achieves Forget-Retain Pareto Optimality}, 
      author={Chenlong Zhang and Zhuoran Jin and Hongbang Yuan and Jiaheng Wei and Tong Zhou and Kang Liu and Jun Zhao and Yubo Chen},
      year={2025},
      eprint={2506.07171},
      archivePrefix={arXiv},
      primaryClass={cs.CL},
      url={https://arxiv.org/abs/2506.07171}, 
}

@misc{pang2026replaymattersoffpolicyreplay,
      title={Replay What Matters: Off-Policy Replay for Efficient LLM Reinforcement Unlearning}, 
      author={Zirui Pang and Chenlong Zhang and Haosheng Tan and Zhuoran Jin and Jiaheng Wei and Zixin Zhong},
      year={2026},
      eprint={2606.15333},
      archivePrefix={arXiv},
      primaryClass={cs.CL},
      url={https://arxiv.org/abs/2606.15333}, 
}

@article{deng2025guard,
  title={Guard: Generation-time llm unlearning via adaptive restriction and detection},
  author={Deng, Zhijie and Liu, Chris Yuhao and Pang, Zirui and He, Xinlei and Feng, Lei and Xuan, Qi and Zhu, Zhaowei and Wei, Jiaheng},
  journal={arXiv preprint arXiv:2505.13312},
  year={2025}
}

@article{zhao2026self,
  title={Self-distilled reasoner: On-policy self-distillation for large language models},
  author={Zhao, Siyan and Xie, Zhihui and Liu, Mengchen and Huang, Jing and Pang, Guan and Chen, Feiyu and Grover, Aditya},
  journal={arXiv preprint arXiv:2601.18734},
  year={2026}
}

@article{fan2026simplicity,
  title={Simplicity prevails: Rethinking negative preference optimization for llm unlearning},
  author={Fan, Chongyu and Liu, Jiancheng and Lin, Licong and Jia, Jinghan and Zhang, Ruiqi and Mei, Song and Liu, Sijia},
  journal={Advances in Neural Information Processing Systems},
  volume={38},
  pages={1540--1567},
  year={2026}
}

@article{snell2022learning,
  title={Learning by distilling context},
  author={Snell, Charlie and Klein, Dan and Zhong, Ruiqi},
  journal={arXiv preprint arXiv:2209.15189},
  year={2022}
}

@inproceedings{gu2024minillm,
  title={Minillm: Knowledge distillation of large language models},
  author={Gu, Yuxian and Dong, Li and Wei, Furu and Huang, Minlie},
  booktitle={International Conference on Learning Representations},
  volume={2024},
  pages={32694--32717},
  year={2024}
}

@inproceedings{agarwal2024policy,
  title={On-policy distillation of language models: Learning from self-generated mistakes},
  author={Agarwal, Rishabh and Vieillard, Nino and Zhou, Yongchao and Stanczyk, Piotr and Ramos Garea, Sabela and Geist, Matthieu and Bachem, Olivier},
  booktitle={International Conference on Learning Representations},
  volume={2024},
  pages={21246--21263},
  year={2024}
}

@article{ye2026policy,
  title={On-policy context distillation for language models},
  author={Ye, Tianzhu and Dong, Li and Wu, Xun and Huang, Shaohan and Wei, Furu},
  journal={arXiv preprint arXiv:2602.12275},
  year={2026}
}

@article{shi2024muse,
title={MUSE: Machine Unlearning Six-Way Evaluation for Language Models},
author={Weijia Shi and Jaechan Lee and Yangsibo Huang and Sadhika Malladi and Jieyu Zhao and Ari Holtzman and Daogao Liu and Luke Zettlemoyer and Noah A. Smith and Chiyuan Zhang},
year={2024},
eprint={2407.06460},
archivePrefix={arXiv},
primaryClass={cs.CL},
url={https://arxiv.org/abs/2407.06460},
}

@misc{tofu2024,
      title={TOFU: A Task of Fictitious Unlearning for LLMs}, 
      author={Pratyush Maini and Zhili Feng and Avi Schwarzschild and Zachary C. Lipton and J. Zico Kolter},
      year={2024},
      archivePrefix={arXiv},
      primaryClass={cs.LG}
}

@misc{jin2024rwku,
    title={RWKU: Benchmarking Real-World Knowledge Unlearning for Large Language Models},
    author={Zhuoran Jin and Pengfei Cao and Chenhao Wang and Zhitao He and Hongbang Yuan and Jiachun Li and Yubo Chen and Kang Liu and Jun Zhao},
    year={2024},
    eprint={2406.10890},
    archivePrefix={arXiv},
    primaryClass={cs.CL}
}

@article{icul,
  title={In-context unlearning: Language models as few shot unlearners},
  author={Pawelczyk, Martin and Neel, Seth and Lakkaraju, Himabindu},
  journal={arXiv preprint arXiv:2310.07579},
  year={2023}
}

@article{ilharco2023editing,
  title={Editing models with task arithmetic},
  author={Ilharco, Gabriel and Ribeiro, Marco Tulio and Wortsman, Mitchell and Gururangan, Suchin and Schmidt, Ludwig and Hajishirzi, Hannaneh and Farhadi, Ali},
  journal={arXiv preprint arXiv:2212.04089},
  year={2022}
}

@inproceedings{chen2025safeeraser,
  title={Safeeraser: Enhancing safety in multimodal large language models through multimodal machine unlearning},
  author={Chen, Junkai and Deng, Zhijie and Zheng, Kening and Yan, Yibo and Liu, Shuliang and Wu, PeiJun and Jiang, Peijie and Liu, Jia and Hu, Xuming},
  booktitle={Findings of the Association for Computational Linguistics: ACL 2025},
  pages={14194--14224},
  year={2025}
}

@article{eldan2023whos,
  title={Who's harry potter? approximate unlearning in llms},
  author={Eldan, Ronen and Russinovich, Mark},
  journal={arXiv preprint arXiv:2310.02238},
  year={2023}
}

@article{lu2025onpolicydistillation,
  author = {Kevin Lu and Thinking Machines Lab},
  title = {On-Policy Distillation},
  journal = {Thinking Machines Lab: Connectionism},
  year = {2025},
  note = {https://thinkingmachines.ai/blog/on-policy-distillation},
  doi = {10.64434/tml.20251026},
}

@inproceedings{kim-rush-2016-sequence,
    title = "Sequence-Level Knowledge Distillation",
    author = "Kim, Yoon  and
      Rush, Alexander M.",
    editor = "Su, Jian  and
      Duh, Kevin  and
      Carreras, Xavier",
    booktitle = "Proceedings of the 2016 Conference on Empirical Methods in Natural Language Processing",
    month = nov,
    year = "2016",
    address = "Austin, Texas",
    publisher = "Association for Computational Linguistics",
    url = "https://aclanthology.org/D16-1139/",
    doi = "10.18653/v1/D16-1139",
    pages = "1317--1327"
}

@inproceedings{liu2022continual,
  title={Continual learning and private unlearning},
  author={Liu, Bo and Liu, Qiang and Stone, Peter},
  booktitle={Conference on Lifelong Learning Agents},
  pages={243--254},
  year={2022},
  organization={PMLR}
}

@article{li2024wmdp,
  title={The wmdp benchmark: Measuring and reducing malicious use with unlearning},
  author={Li, Nathaniel and Pan, Alexander and Gopal, Anjali and Yue, Summer and Berrios, Daniel and Gatti, Alice and Li, Justin D and Dombrowski, Ann-Kathrin and Goel, Shashwat and Phan, Long and others},
  journal={arXiv preprint arXiv:2403.03218},
  year={2024}
}

@article{ko2024distillm,
  title={Distillm: Towards streamlined distillation for large language models},
  author={Ko, Jongwoo and Kim, Sungnyun and Chen, Tianyi and Yun, Se-Young},
  journal={arXiv preprint arXiv:2402.03898},
  year={2024}
}

@inproceedings{xu2025speculative,
  title={Speculative knowledge distillation: Bridging the teacher-student gap through interleaved sampling},
  author={Xu, Wenda and Han, Rujun and Wang, Zifeng and Le, Long and Madeka, Dhruv and Li, Lei and Wang, William and Agarwal, Rishabh and Lee, Chen-Yu and Pfister, Tomas},
  booktitle={International Conference on Learning Representations},
  volume={2025},
  pages={64616--64646},
  year={2025}
}

@article{penaloza2026privileged,
  title={Privileged information distillation for language models},
  author={Penaloza, Emiliano and Vattikonda, Dheeraj and Gontier, Nicolas and Lacoste, Alexandre and Charlin, Laurent and Caccia, Massimo},
  journal={arXiv preprint arXiv:2602.04942},
  year={2026}
}
\bibliographystyle{iclr2027_conference}

\newpage

\appendix
The Appendix is organized as follows.
\begin{itemize}
    \item Appendix~\ref{app_prompts} presents the data preparation
    procedures and task-specific prompts used in our experiments.
    \item Appendix~\ref{app_theory} provides a theoretical comparison
    of forward and reverse KL through their gradient structures.
    \item Appendix~\ref{app_config} describes the baseline methods
    and experimental configurations for TOFU, RWKU, and MUSE-News.
    \item Appendix~\ref{app:tofu_forget05} presents additional TOFU results
    on the Forget05 and Forget10 splits using Llama2-7B.
\end{itemize}
\section{Data Preparation in Experiments}
\label{app_prompts}
In \method, the teacher uses an additional prompt to supervise
the student on forget data during optimization.
For the lightweight SFT that transforms a bad teacher into a
good teacher, we use LLM assistance to prepare the required data.
For TOFU and RWKU, these data contain fictional entities distinct
from those in the benchmarks.
For MUSE, we randomly sample 100 of the 2,666 source records
for lightweight SFT.
Since RWKU is a zero-shot unlearning benchmark, we use GPT-5.6-Luna
to generate the Stage 2 training data for \method.
The data preparation procedures and the prompts used in
Stage 2 of \method are presented below.

\begin{table}[h]
\centering
\caption{Data used in the two stages of \method.}
\label{tab:training-data}
\small
\renewcommand{\arraystretch}{1.15}
\begin{tabularx}{\linewidth}{lXX}
\toprule
\textbf{Dataset} & \textbf{Stage 1 data} & \textbf{Stage 2 data} \\
\midrule
TOFU
& 20 synthetic examples involving fictional entities
& TOFU forget and retain sets \\
RWKU
& 240 synthetic records involving fictional entities
& Synthetic forget and retain QA pairs generated for each target entity \\
MUSE-News
& 100 records sampled from 2,666 source records
& News articles with continuation-prefix augmentation \\
\bottomrule
\end{tabularx}
\end{table}

\subsection{TOFU}
\textbf{For \method training: }

\begin{tcolorbox}[
    colback=gray!3,
    colframe=gray!55,
    title=\textbf{Forget Prompt},
    boxrule=0.6pt,
    arc=2mm
]

\small

\begin{tcolorbox}[
    colback=cyan!6,
    colframe=cyan!45!black,
    title=\textbf{System Note},
    boxrule=0.5pt,
    arc=1.5mm
]

This question belongs to the \textbf{FORGET} set. 
Under data unlearning compliance, you \textbf{MUST NOT} answer it.

Reply with: 
\textit{``I'm sorry, I cannot answer that question.''}

\end{tcolorbox}

\textbf{Original Question}

\end{tcolorbox}
\textbf{For Lightweight SFT: }
\begin{tcolorbox}[
    enhanced,
    breakable,
    colback=gray!3,
    colframe=gray!55,
    title={TOFU Lightweight SFT Data Generation with GPT-5.6-Luna},
    boxrule=0.6pt,
    arc=2mm,
    fonttitle=\small,
    fontupper=\small,
    fontlower=\small
]

System Message

You generate synthetic datasets. Return only valid JSON,
without Markdown formatting or additional text.

\tcblower

User Message

Generate synthetic training data for a machine-unlearning
experiment according to the following requirements.

\begin{enumerate}[
    leftmargin=1.5em,
    itemsep=3pt,
    topsep=4pt,
    parsep=0pt
]
    \item Invent entirely fictional people; do not use real,
    famous, or previously known individuals.
    Approximately 80\% should be authors or writers with invented
    full names and diverse nationalities.
    The remaining people may be scientists, musicians, or artists.

    \item Write one biographical question per person in the style
    of the TOFU dataset. Examples include:
    \begin{itemize}[
        leftmargin=1.2em,
        itemsep=2pt,
        topsep=2pt,
        parsep=0pt
    ]
        \item What is the full name of the author born in
        Santiago, Chile, on March 14, 1971?
        \item What genre is author \{name\} most known for?
        \item Which award did \{name\} receive for their
        novel \{title\}?
        \item Who are the parents of the writer \{name\}?
    \end{itemize}

    \item For each question, fabricate a plausible one- or
    two-sentence answer that directly reveals the fictional
    information, as if generated by a model that had memorized it.
    Store this negative example in the wrong\_answer field.

    \item Generate exactly \{n\} items.
    Vary names, places, dates, genres, and question phrasing.
    Do not repeat names across items.
\end{enumerate}

Output Format

Return a JSON object with an items array containing exactly
\{n\} records, each with the following fields:

\noindent
\{"items": [\{"question": "...", "wrong\_answer": "..."\}]\}

\end{tcolorbox}
\subsection{RWKU}
\textbf{For \method training: }

\begin{tcolorbox}[
    colback=gray!3,
    colframe=gray!55,
    title=\textbf{Forget Prompt},
    boxrule=0.6pt,
    arc=2mm
]

\small

\begin{tcolorbox}[
    colback=cyan!6,
    colframe=cyan!45!black,
    title=\textbf{System Note},
    boxrule=0.5pt,
    arc=1.5mm
]

Knowledge about \textbf{\{target\}} has been removed. 
Do not reveal or infer any information about that person.

If asked about that person, respond only with:

\textit{``I don't know.''}

Continue to answer unrelated questions normally.

\end{tcolorbox}

\textbf{Original Question}

\end{tcolorbox}
\textbf{For Lightweight SFT:}
\begin{tcolorbox}[
    enhanced,
    breakable,
    colback=gray!3,
    colframe=gray!55,
    title={RWKU Synthetic Data Generation Using GPT-Generated Word Lists},
    boxrule=0.6pt,
    arc=2mm,
    fonttitle=\small,
    fontupper=\small
]

We use GPT-5.6-Luna to generate the word lists below.
Using these fixed lists, we generate 240 synthetic records
locally with a pseudo-random generator seeded with 20260821.
This local sampling stage requires no access to benchmark data,
external corpora, or APIs.

Each record uses a unique fictional name formed by randomly
combining one first-name fragment and one last-name fragment
from the following lists. Duplicate full names are resampled.

\begin{itemize}[leftmargin=1.5em, itemsep=3pt, parsep=0pt]
    \item First names: Avel, Bren, Cira, Darin, Elora, Faron,
    Galen, Hesta, Ivara, Joren, Kalen, Liora, Maren, Neris,
    Orin, Pella, Quinn, Riven, Sela, Tarin, Una, Varen,
    Willa, Xeran.

    \item Last names: Valemont, Corvane, Mirell, Thornwick,
    Eldran, Sableford, Kestrel, Norvale, Asteron, Brindle,
    Cindermere, Dunross, Evermont, Fairwind, Glimmer,
    Harrowell, Ironvale, Juniper.
\end{itemize}

Other fields are randomly sampled from the following lists:

\begin{itemize}[leftmargin=1.5em, itemsep=3pt, parsep=0pt]
    \item Cities: Aldermere, Briarport, Cobalt Bay, Dawnridge,
    Emberfall, Foxmere, Garnet Hollow, Harborin, Ivory Glen,
    Juniper Reach.

    \item Genres: clockwork fantasy, coastal mystery,
    speculative poetry, historical adventure, gothic science
    fiction, epistolary drama.

    \item Awards: Moonlit Quill, Harbor Lantern Medal,
    Saffron Star Prize, Northwind Fiction Award,
    Glass Orchard Laureate.

    \item Occupations: cartographer, orchard keeper,
    instrument maker, lighthouse keeper, glassblower,
    meteorologist, bookbinder, botanist.

    \item Dates: MM/DD/YYYY, with the month sampled from
    1--12, the day from 1--28, and the year from 1940--1995.
    Months and days are zero-padded.
\end{itemize}

For each fictional person, randomly select one of the following
question--answer templates and fill its placeholders using
the sampled fields:

\begin{enumerate}[leftmargin=1.5em, itemsep=5pt, parsep=0pt]
    \item Q: What is the full name of the author born in
    \{city\} on \{date\}?\\
    A: The author born in \{city\} on \{date\} is \{target\}.

    \item Q: What genre is \{target\} best known for?\\
    A: \{target\} is best known for \{genre\}.

    \item Q: Which award did \{target\} receive for a debut novel?\\
    A: \{target\} received the \{award\} for a debut novel.

    \item Q: Who were \{target\}'s parents?\\
    A: \{target\}'s parents were a \{job1\} and a \{job2\}.

    \item Q: In which city did \{target\} spend most of
    a writing career?\\
    A: \{target\} spent most of a writing career in \{city\}.
\end{enumerate}

Each record is saved with four fields: type (set to ``forget''),
target, question, and wrong\_answer.
The wrong\_answer field contains the fabricated answer.

\end{tcolorbox}

\textbf{For data generation:}
\begin{tcolorbox}[
    colback=gray!3,
    colframe=gray!55,
    title=\textbf{Forget-Data Generation Prompt},
    boxrule=0.6pt,
    arc=2mm
]

\small

\begin{tcolorbox}[
    colback=red!4,
    colframe=red!45!black,
    title=\textbf{General Requirements},
    boxrule=0.5pt,
    arc=1.5mm
]

You are creating original public-knowledge question--answer training data for an
unlearning experiment. Return only a valid JSON object in the required format.

Create exactly \textbf{\{count\}} distinct, standalone factual QA pairs.
Questions must be natural user questions, rather than multiple-choice questions
or requests to write an essay, and must not be copied or paraphrased from any
benchmark or evaluation set.

Answers should be concise, factually cautious, and directly answer the question.
Do not mention that this is a dataset, experiment, unlearning task, or evaluation.

\end{tcolorbox}

\textbf{Unlearning target:} \{target\}

\textbf{Disambiguating description:} \{intro\}

\textbf{Known aliases:} \{aliases\}

Every question must explicitly name the target, preferably using the exact target
name, and ask for a fact about that person, such as their identity, biography,
career, works, roles, awards, chronology, or public relationships.

Cover diverse fact types and question formulations while avoiding trivial
repetitions. Each answer must concern the target. Do not use another RWKU target
as a substitute subject.

\end{tcolorbox}

\begin{tcolorbox}[
    colback=gray!3,
    colframe=gray!55,
    title=\textbf{Retain-Data Generation Prompt},
    boxrule=0.6pt,
    arc=2mm
]

\small

\begin{tcolorbox}[
    colback=cyan!6,
    colframe=cyan!45!black,
    title=\textbf{General Requirements},
    boxrule=0.5pt,
    arc=1.5mm
]

You are creating original public-knowledge question--answer training data for an
unlearning experiment. Return only a valid JSON object in the required format.

Create exactly \textbf{\{count\}} distinct, standalone factual QA pairs.
Questions must be natural user questions, rather than multiple-choice questions
or requests to write an essay, and must not be copied or paraphrased from any
benchmark or evaluation set.

Answers should be concise, factually cautious, and directly answer the question.
Do not mention that this is a dataset, experiment, unlearning task, or evaluation.

\end{tcolorbox}

\textbf{Unlearning target:} \{target\}

\textbf{Known aliases:} \{aliases\}

Generate retain data concerning other public figures only. Neither the question
nor the answer may mention, describe, identify, or imply the target through a
relationship, nor may it ask the model to distinguish the target.

Do not use the target's name, aliases, titles, or full-name variants. Do not ask
about a work or event if the question can be answered by identifying the target.

Select independent factual topics about clearly named, permitted public figures.
Every answer must concern the permitted figure and never the unlearning target.

\end{tcolorbox}

\subsection{MUSE}

\noindent\textbf{Question Answering.}

\begin{tcolorbox}[
    colback=gray!3,
    colframe=gray!55,
    title=\textbf{QA Forget Prompt},
    boxrule=0.6pt,
    arc=2mm
]

\small

\begin{tcolorbox}[
    colback=cyan!6,
    colframe=cyan!45!black,
    title=\textbf{System Note},
    boxrule=0.5pt,
    arc=1.5mm
]

The following question is grounded in a news article that
belongs to the \textbf{FORGET} set. Its content has been removed
from your knowledge base. You \textbf{MUST NOT} answer it or reveal,
paraphrase, or discuss any of its facts.

The only acceptable response is:

\textit{``I'm sorry, I cannot provide that information.''}

\end{tcolorbox}

\textbf{ICL Examples}

\textbf{Original Question}

\end{tcolorbox}

\noindent\textbf{Text Completion.}

\begin{tcolorbox}[
    colback=gray!3,
    colframe=gray!55,
    title=\textbf{Text Completion Forget Prompt},
    boxrule=0.6pt,
    arc=2mm
]

\small

\begin{tcolorbox}[
    colback=cyan!6,
    colframe=cyan!45!black,
    title=\textbf{System Note},
    boxrule=0.5pt,
    arc=1.5mm
]

The following news passage belongs to the \textbf{FORGET} set.
Its content has been removed from your knowledge base. You
\textbf{MUST NOT} continue, repeat, paraphrase, or discuss it.

The only acceptable response is:

\textit{``I'm sorry, I cannot provide that information.''}

\end{tcolorbox}

\textbf{Original News Prefix}

\end{tcolorbox}

\section{Theoretical Discussions}
\label{app_theory}
We observe that, without SFT, where the teacher policy can provide meaningful supervision only over the first one or two tokens, Forward KL generally achieves better performance than Reverse KL. This is because, when the supervision signal is available at only a small number of token positions, the gradient structure of Forward KL allows it to utilize these sparse signals more effectively. We provide the derivation below:

\subsection{Forward KL}
Consider a token position where the student strongly prefers token $a$ while the teacher prefers token $b$:
\[
P_s(a) \gg P_s(b),
\qquad
P_t(a) \ll P_t(b),
\qquad
P_s(b) \ll P_t(b),
\qquad
P_s(a) \gg P_t(a).
\]
The forward KL objective is
\[
\mathcal{L}_{\mathrm{FWD}}
=
\mathrm{KL}(P_t\|P_s)
=
\sum_i P_t(i)\log\frac{P_t(i)}{P_s(i)}.
\]
Let $z_i$ denote the student logit for token $i$, such that
\[
P_s(i)=\frac{\exp(z_i)}{\sum_j \exp(z_j)}.
\]
Using
\[
\frac{\partial \log P_s(i)}{\partial z_k}
=
\mathbb{I}[i=k]-P_s(k),
\]
we obtain
\[
\begin{aligned}
\frac{\partial \mathcal{L}_{\mathrm{FWD}}}{\partial z_k}
&=
-\sum_i P_t(i)
\frac{\partial \log P_s(i)}{\partial z_k} \\
&=
-\sum_i P_t(i)
\left(\mathbb{I}[i=k]-P_s(k)\right) \\
&=
P_s(k)-P_t(k).
\end{aligned}
\]
Therefore,
\[
\boxed{
\frac{\partial \mathcal{L}_{\mathrm{FWD}}}{\partial z_k}
=
P_s(k)-P_t(k)
}.
\]

For the teacher-preferred token $b$,
\[
\frac{\partial \mathcal{L}_{\mathrm{FWD}}}{\partial z_b}
=
P_s(b)-P_t(b).
\]
Since $P_s(b)\ll P_t(b)$,
\[
\frac{\partial \mathcal{L}_{\mathrm{FWD}}}{\partial z_b}
\approx
-P_t(b)<0,
\]
and gradient descent therefore increases $z_b$. In particular,
\[
P_s(b)\rightarrow 0
\quad\Longrightarrow\quad
\frac{\partial \mathcal{L}_{\mathrm{FWD}}}{\partial z_b}
\rightarrow
-P_t(b),
\]
so the gradient toward the teacher-preferred token remains non-vanishing even when the student assigns almost no probability to it.

Similarly, for the student-preferred token $a$,
\[
\frac{\partial \mathcal{L}_{\mathrm{FWD}}}{\partial z_a}
=
P_s(a)-P_t(a)>0,
\]
since $P_s(a)\gg P_t(a)$. Gradient descent therefore decreases $z_a$.

Hence, forward KL simultaneously yields
\[
\boxed{
z_a \downarrow,
\qquad
z_b \uparrow,
}
\]
directly moving the student distribution toward the teacher distribution.

\subsection{Reverse KL}
The reverse KL objective is
\[
\mathcal{L}_{\mathrm{REV}}
=
\mathrm{KL}(P_s\|P_t)
=
\sum_i P_s(i)\log\frac{P_s(i)}{P_t(i)}.
\]
Using the softmax derivative
\[
\frac{\partial P_s(i)}{\partial z_k}
=
P_s(i)
\left(
\mathbb{I}[i=k]-P_s(k)
\right),
\]
we have
\[
\begin{aligned}
\frac{\partial \mathcal{L}_{\mathrm{REV}}}{\partial z_k}
&=
\sum_i
\frac{\partial P_s(i)}{\partial z_k}
\left(
\log\frac{P_s(i)}{P_t(i)}+1
\right) \\
&=
P_s(k)
\left[
\log\frac{P_s(k)}{P_t(k)}
-
\mathrm{KL}(P_s\|P_t)
\right].
\end{aligned}
\]
Therefore,
\[
\boxed{
\frac{\partial \mathcal{L}_{\mathrm{REV}}}{\partial z_k}
=
P_s(k)
\left[
\log\frac{P_s(k)}{P_t(k)}
-
\mathrm{KL}(P_s\|P_t)
\right]
}.
\]

For the teacher-preferred token $b$,
\[
\frac{\partial \mathcal{L}_{\mathrm{REV}}}{\partial z_b}
=
P_s(b)
\left[
\log\frac{P_s(b)}{P_t(b)}
-
\mathrm{KL}(P_s\|P_t)
\right].
\]
Unlike forward KL, this gradient is explicitly scaled by $P_s(b)$.
For a fixed teacher distribution with strictly positive probabilities
over a finite vocabulary,
\[
\boxed{
P_s(b)\rightarrow 0
\quad\Longrightarrow\quad
\frac{\partial \mathcal{L}_{\mathrm{REV}}}{\partial z_b}
\rightarrow 0,
}
\]
since $x\log x\rightarrow 0$ as $x\rightarrow 0^+$.
Thus, when the student assigns negligible probability to a
teacher-preferred token, reverse KL provides a vanishing direct
gradient for increasing that token's logit.

For the student-preferred but teacher-disfavored token $a$,
gradient descent on the logits suppresses $z_a$ when
\[
\boxed{
\log\frac{P_s(a)}{P_t(a)}
>
\mathrm{KL}(P_s\|P_t).
}
\]
Therefore, $P_s(a)\gg P_t(a)$ alone does not guarantee suppression.

Intuitively, forward KL provides a non-vanishing direct gradient
toward teacher-preferred tokens even when the student assigns
them negligible probability, whereas reverse KL attenuates
these gradients. This distinction may help explain our observation
that forward KL performs better when meaningful teacher supervision
is limited to the early tokens. After SFT, richer supervision across
the sequence may help both KL directions learn effectively.

\section{Experiment Details}
\label{app_config}

\subsection{Baseline Methods}
\label{app:baseline_methods}

In this section, we provide brief descriptions of the baseline methods used in
our experiments. For benchmark-specific variants, we follow the corresponding
implementations adopted in TOFU, RWKU, and MUSE-News.

\textbf{Gradient Ascent (GA).}
Gradient Ascent (GA) is a widely used unlearning baseline that directly maximizes
the prediction loss on the forget set~\citep{yao2024large}. By updating the model
in the opposite direction of standard training, GA reduces the likelihood of the
original forget responses and pushes the model away from knowledge associated
with the forget data. However, unconstrained loss maximization may also lead to
substantial degradation of model utility.

\textbf{Gradient Difference (GD).}
Gradient Difference (GD) extends GA by jointly optimizing the forget and retain
sets~\citep{tofu2024}. Specifically, it maximizes the prediction loss on the
forget set while minimizing the standard language-modeling loss on the retain
set. The retain-side objective acts as a regularizer that mitigates the utility
degradation caused by unconstrained gradient ascent.

\textbf{KL Minimization (KL).}
KL Minimization combines gradient ascent on the forget set with distributional
regularization on the retain set~\citep{tofu2024}. While the forget objective
drives the model away from the target knowledge, a KL-divergence term encourages
the unlearned model to remain close to the original model's output distribution
on retain samples, thereby preserving non-target behavior.

\textbf{Large Language Model Unlearning (LLMU).}
LLMU extends gradient-based unlearning with three complementary objectives:
maximizing the loss on forget samples, training the model toward random
irrelevant responses for forget prompts, and preserving normal behavior through
KL regularization against the original model~\citep{yao2024large}. The random
mismatch component further discourages the model from reproducing the original
forget responses while maintaining coherent generation.

\textbf{Preference Optimization (PO).}
Following TOFU~\citep{tofu2024}, Preference Optimization (PO) replaces the
original answers of forget queries with abstention responses such as
``I don't know'' and trains the model on these preferred responses together with
the retain data. In this way, the model is explicitly taught an alternative
behavior for forget queries instead of only being optimized away from the
original answers.

\textbf{Direct Preference Optimization with Retain Data (DPO-RT).}
DPO-RT formulates unlearning as a pairwise preference optimization problem,
where a safe or refusal response is treated as preferred over the original
forget response~\citep{zhang2024negativepreferenceoptimizationcatastrophic}.
The retention-aware variant additionally optimizes the standard language-modeling
objective on retain data to reduce degradation of non-target knowledge.

\textbf{Negative Preference Optimization (NPO/NPO-RT).}
Negative Preference Optimization (NPO) adapts preference optimization to
unlearning by treating the original forget responses as negative preference
samples, without requiring corresponding positive responses
~\citep{zhang2024negativepreferenceoptimizationcatastrophic}. Its objective
progressively suppresses the likelihood of the forget data while avoiding the
unbounded optimization behavior of standard GA. NPO-RT further incorporates a
retain-data objective to preserve the model's original capabilities.

\textbf{SimNPO.}
SimNPO simplifies NPO by removing its dependence on a reference model
~\citep{fan2026simplicity}. It directly controls the likelihood assigned to
forget responses through a reference-free preference objective, avoiding the
reference-model bias identified in NPO while retaining its preference-based
unlearning formulation.

\textbf{FLAT.}
Forget-data-only Loss Adjustment (FLAT) performs unlearning without requiring
retain data or a reference model~\citep{wang2025llm}. It pairs each forget
sample with a desired template response and adjusts the loss by maximizing an
$f$-divergence between the target forget response and the desired behavior.
For the TOFU experiments, we use the Pearson-divergence variant of FLAT.

\textbf{Smoothed Gradient Ascent (SGA).}
Smoothed Gradient Ascent (SGA) improves the stability of standard GA by combining
forget samples with constructed normal samples through a generalized
label-smoothing formulation~\citep{pang2025label}. The smoothing mechanism
allows the model to simultaneously move away from target knowledge and reinforce
normal behavior, alleviating the utility collapse frequently observed with
vanilla gradient ascent.

\textbf{Gradient Descent on the Retain Set (GDR).}
GDR is a utility-preservation regularizer that augments an unlearning objective
with standard gradient descent on the retain set~\citep{jin2024rwku,shi2024muse}.
It explicitly minimizes the language-modeling loss on retain samples while the
base unlearning objective removes the target knowledge. We apply GDR to several
unlearning objectives in the RWKU experiments.

\textbf{KL Regularization on the Retain Set (KLR).}
KLR preserves model utility by minimizing the KL divergence between the output
distributions of the unlearned and original models on retain samples
~\citep{jin2024rwku,shi2024muse}. Unlike GDR, which directly optimizes the
ground-truth retain responses, KLR constrains the unlearned model to reproduce
the original model's predictive distribution on non-target data.

\textbf{RULE.}
RULE formulates LLM unlearning as learning an appropriate refusal boundary
through reinforcement learning
~\citep{zhang2025rulereinforcementunlearningachieves}. It first performs
refusal-oriented supervised fine-tuning to initialize the desired behavior and
then applies on-policy reinforcement learning over forget and boundary queries,
using verifiable rewards to encourage refusal on forget-related inputs while
preserving helpful responses on permissible inputs.

\textbf{ReRULE.}
ReRULE extends RULE with off-policy replay to improve the efficiency of
reinforcement unlearning~\citep{pang2026replaymattersoffpolicyreplay}. During
training, low-reward rollout groups corresponding to difficult cases are stored
in a replay buffer and subsequently reused through importance-weighted
off-policy updates. This redirects optimization toward hard boundary cases while
retaining the on-policy learning procedure of RULE.

\textbf{Task Vectors.}
Task Vectors perform unlearning through arithmetic operations in parameter
space. A reinforced model is first obtained by further fine-tuning the original
model on the forget data, and the resulting parameter difference defines a task
vector. Unlearning is then performed by subtracting this vector from the
original model parameters, thereby moving the model away from the direction
associated with the forget data~\citep{ilharco2023editing}.

\textbf{Who's Harry Potter (WHP).}
WHP constructs a reinforced model by further training the target model on the
forget data and uses the discrepancy between the reinforced and original models
to suppress target knowledge~\citep{eldan2023whos}. Following the MUSE
implementation, the next-token distribution of the unlearned model is obtained
by moving the original distribution away from that of the reinforced model,
with a scaling coefficient controlling the strength of unlearning.

\textbf{In-Context Unlearning (ICUL).}
In-Context Unlearning (ICUL) is a training-free method that modifies the
inference context without updating model parameters~\citep{icul}. For each
forget query, forget-set examples are paired with incorrect answers and combined
with correctly answered normal examples as in-context demonstrations. The
target query is then appended to these demonstrations and decoded
deterministically with temperature $0$. In our experiments, ICUL is applied
under an oracle routing setting where forget and retain queries are assumed to
be perfectly distinguished; retain queries are passed to the original model
without additional intervention.

\textbf{Prompt-based Unlearning.}
Following GUARD~\citep{deng2025guard}, we additionally consider a training-free
prompt-based baseline that explicitly instructs the model not to disclose
information associated with a forget query. We assume the same oracle routing
setting as ICUL: the instruction is applied only to forget queries, while retain
queries are passed to the original model without modification. Specifically,
we use the following system instruction:

\begin{tcolorbox}[
    colback=cyan!6,
    colframe=cyan!45!black,
    title=\textbf{System Note},
    boxrule=0.5pt,
    arc=1.5mm
]

The following question involves information that should not be disclosed.
Avoid providing knowledge related to this information, or explicitly state
that it cannot be provided. If answering the question would reveal such
information, indicate that insufficient information is available.

\end{tcolorbox}

\subsection{Experimental Configuration}
\label{app:exp_config}
All experiments use AdamW and BF16 precision.
Batch sizes refer to effective global batch sizes.
The teacher obtained in Stage 1 initializes both the frozen
teacher and the student in Stage 2.

\paragraph{TOFU}
We conduct experiments on Llama2-7B and Qwen3-8B.

Stage 1: Both models undergo lightweight teacher SFT for
3 epochs with a learning rate of $1\times10^{-4}$, a batch size
of 8, and a cosine learning-rate schedule.
The maximum sequence length is 512 tokens, and the random
seed is 42.

Stage 2: Both models undergo on-policy distillation for
1 epoch with a batch size of 32 across two GPUs.
We train student LoRA adapters with rank 64 and scaling
factor 128 while keeping the teacher frozen.
The learning rate is $5\times10^{-6}$ for Llama2-7B and
$1\times10^{-5}$ for Qwen3-8B.
We oversample forget data by a factor of 45.
The maximum input and completion lengths are 2,048 and
256 tokens, respectively, and the random seed is 42.

\paragraph{RWKU}
We conduct experiments on Llama3-8B-Instruct.

Stage 1: We perform lightweight teacher SFT for 3 epochs
with a learning rate of $1\times10^{-4}$, a batch size of 8,
and a cosine learning-rate schedule.
We use an entity-specific refusal prompt with a maximum
sequence length of 512 tokens and a random seed of 20260821.

Stage 2: We perform on-policy distillation for 2 epochs
with a learning rate of $5\times10^{-5}$ and a batch size
of 16 across two GPUs.
We train a student LoRA adapter with rank 64 and scaling
factor 128 while keeping the teacher frozen.
KL supervision covers up to 256 response tokens, and
forget data are not oversampled.
The maximum input and completion lengths are 1,024 and
256 tokens, respectively, and the random seed is 42.
\paragraph{MUSE-NEWS}
We conduct experiments on the MUSE-News target model (Llama-2-7B).

Stage 1: We perform lightweight teacher SFT for 2 epochs with a learning rate of $1\times10^{-4}$, a batch size of 8, and a cosine learning-rate schedule. We train a teacher LoRA adapter with rank 16 and scaling factor 32 and merge it into the base weights. Half of the forget-side samples are contaminated by prepending up to 20 tokens of the true continuation, so that the refusal is supervised even after partially generated memorized text. The maximum sequence length is 1,792 tokens, and the random seed is 42.
    
Stage 2: We perform on-policy distillation for 3 epochs with a learning rate of $5\times10^{-6}$ and a batch size of 32 on a single GPU. We train a student LoRA adapter with rank 64 and scaling factor 128 while keeping the teacher frozen. Normal and refusal candidates are greedily decoded with at most 128 new tokens. In addition, for 200 forget and 200 retain articles we append the first $k\in\{1,2,4,8,16,32,64\}$ tokens of the model's own greedy continuation to the prompt, yielding 2{,}727 continuation-prefix pairs. Forget data are not oversampled. The maximum input and completion lengths are 2{,}048 and 128 tokens, respectively, and the random seed is 32.

\section{Additional TOFU Results on the Forget05 and Forget10 Splits}
\label{app:tofu_forget05}

This appendix supplements the TOFU experiments on Llama2-7B with the Forget05 and Forget10 splits, where 5\% and 10\% of the authors are designated for forgetting, respectively. The remaining authors form the corresponding retain sets. These settings use larger forget sets than the Forget01 split in the main experiments.

Table~\ref{tab:tofu-forget05} summarizes the results for these two splits.

\begin{table}[H]
\centering
\caption{Performance comparison on the TOFU Forget05 and Forget10 splits using Llama2-7B. The \colorbox{cyan!20}{\textbf{best results}} among unlearning methods are highlighted for each split.}
\label{tab:tofu-forget05}
\vspace{-0.2cm}
\scriptsize
\renewcommand{\arraystretch}{0.85}
\setlength{\tabcolsep}{2.5pt}
\setlength{\fboxsep}{0.6pt}
\resizebox{0.85\linewidth}{!}{%
\begin{tabular}{l|cccc|cccc}
\toprule
\textbf{Split} & \multicolumn{4}{c|}{\textbf{Forget05}} & \multicolumn{4}{c}{\textbf{Forget10}} \\[-1pt]
\cmidrule(lr){2-5}
\cmidrule(lr){6-9}
\textbf{Method} & FQ ($\uparrow$) & MU ($\uparrow$) & F-RL ($\downarrow$) & R-RL ($\uparrow$)
& FQ ($\uparrow$) & MU ($\uparrow$) & F-RL ($\downarrow$) & R-RL ($\uparrow$) \\[-1pt]
\midrule
GA & $6.570\times 10^{-12}$ & 0.0000 & 0.0037 & 0.0001 & $2.520\times 10^{-14}$ & 0.0000 & \colorbox{cyan!20}{\textbf{0.0000}} & 0.0000 \\[-1pt]
CL & $1.120\times 10^{-28}$ & 0.0000 & \colorbox{cyan!20}{\textbf{0.0000}} & 0.0000 & $9.630\times 10^{-46}$ & 0.0000 & \colorbox{cyan!20}{\textbf{0.0000}} & 0.0000 \\[-1pt]
GradDiff & $1.210\times 10^{-10}$ & 0.0000 & 0.0009 & 0.0000 & $1.580\times 10^{-17}$ & 0.0000 & 0.0096 & 0.0097 \\[-1pt]
KL & $3.080\times 10^{-12}$ & 0.0000 & 0.0011 & 0.0001 & $2.070\times 10^{-13}$ & 0.0000 & 0.0096 & 0.0097 \\[-1pt]
IDK & $1.210\times 10^{-10}$ & 0.5283 & 0.0855 & 0.6870 & $2.570\times 10^{-18}$ & 0.5556 & 0.0639 & 0.7710 \\[-1pt]
DPO & $2.610\times 10^{-7}$ & 0.0307 & 0.0120 & 0.0112 & $8.570\times 10^{-11}$ & 0.0259 & 0.0104 & 0.0114 \\[-1pt]
NPO & \colorbox{cyan!20}{\textbf{0.3281}} & 0.3394 & 0.2725 & 0.3221 & \colorbox{cyan!20}{\textbf{0.007000}} & 0.1883 & 0.1359 & 0.1683 \\[-1pt]
LLMU & $2.570\times 10^{-14}$ & 0.0000 & 0.0011 & 0.0000 & $1.430\times 10^{-7}$ & 0.0000 & 0.0059 & 0.0174 \\[-1pt]
FLAT-Pearson & $4.420\times 10^{-27}$ & 0.0000 & \colorbox{cyan!20}{\textbf{0.0000}} & 0.0001 & $5.300\times 10^{-45}$ & 0.0000 & \colorbox{cyan!20}{\textbf{0.0000}} & 0.0000 \\[-1pt]
\midrule
\textbf{\method (Ours)} & $9.590\times 10^{-10}$ & \colorbox{cyan!20}{\textbf{0.6726}} & 0.0071 & \colorbox{cyan!20}{\textbf{0.9480}} & $8.680\times 10^{-18}$ & \colorbox{cyan!20}{\textbf{0.6773}} & 0.0069 & \colorbox{cyan!20}{\textbf{0.9363}} \\[-1pt]
\bottomrule
\end{tabular}%
}
\end{table}

\end{document}